\documentclass{elsarticle}
\usepackage{amssymb}
\usepackage{rotating}
\usepackage{paralist}
\usepackage{graphicx}
\usepackage{tabularx}
\usepackage{siunitx}
\usepackage[unicode]{hyperref}
\usepackage{fancyvrb}
\usepackage{url}
\usepackage{amsmath}
\usepackage{booktabs}
\usepackage{ntheorem}
\usepackage{bm}
\usepackage{amssymb}
\usepackage{blindtext}
\usepackage{tikz}
\usepackage{lineno,hyperref}
\usepackage{subcaption}
\usepackage{booktabs}
\usepackage{framed}
\usepackage{etoolbox}  
\newcommand\eatpunct[1]{}
\usepackage{paralist}
\AtBeginEnvironment{algorithmic}{\footnotesize} 

\usepackage{tikz}

\AtBeginEnvironment{tabular}{\footnotesize} 
\usepackage{algorithm,algcompatible}
\newcommand{\mybox}[1]{%
         \begin{center}%
            \begin{tikzpicture}%
                \node[rectangle, draw=gray, top color=white!10, bottom color=white!90, rounded corners=5pt, inner xsep=5pt, inner ysep=6pt, outer ysep=10pt]{
                \begin{minipage}{0.95\linewidth}#1\end{minipage}};%
            \end{tikzpicture}%
         \end{center}%
}

\usepackage{todonotes}
\begin{document}
\begin{frontmatter}
\title{A review of possible effects of cognitive biases on interpretation of rule-based machine learning models}
\author[A]{Tom{\'a}\v{s} Kliegr}\corref{cor1}
\author[B]{\v{S}t\v{e}p{\'a}n Bahn{\'i}k}
\author[C]{Johannes F\"urnkranz}
\address[A]{Department of Information and Knowledge Engineering, \\ Prague University of Economics and Business,  Czech Republic\\
E-mail: tomas.kliegr@vse.cz}
\address[B]{The Prague College of Psychosocial Studies, Czech Republic\\
E-mail: bahniks@seznam.cz}
\address[C]{
    Johannes Kepler University,
    Computational Data Analytics Group, Linz, Austria\\
    Email: juffi@faw.jku.at\\}
\cortext[cor1]{Corresponding author}

\begin{abstract}
While the interpretability of machine learning models is often equated with their 
mere syntactic comprehensibility, we think that interpretability goes beyond that, and that human interpretability should also be investigated from the point of view of cognitive science. In particular, the goal of this paper is to 
discuss to what extent cognitive biases may affect human understanding of
interpretable machine learning models, in particular of logical rules discovered from data. Twenty
cognitive biases are covered, as are 
possible debiasing techniques that can be adopted by designers of machine learning algorithms and software. 
Our review transfers results obtained in cognitive psychology to the domain of machine learning, aiming to bridge the current gap between these two areas. It needs to be followed by empirical studies specifically focused on the machine learning domain.
\end{abstract}


\end{frontmatter}

\section{Introduction}

This paper aims to investigate the possible effects of cognitive biases on human understanding of machine learning models, in particular inductively learned rules. We use the term ``cognitive bias'' as a representative for various cognitive phenomena that materialize themselves in the form of occasionally irrational reasoning patterns, which are thought to allow humans to make fast judgments and decisions.

Their cumulative effect on human reasoning should not be underestimated as 
``cognitive biases seem  reliable, systematic, and difficult to eliminate'' \citep{kahneman1972subjective}. The effect of some cognitive biases is more pronounced when people do not have well-articulated preferences \citep{tversky1993context}, which is often the case in explorative data analysis.

Previous works have analysed the impact of cognitive biases on multiple types of human behaviour and decision making. A specific example is the seminal book ``Social cognition'' by \citet{kunda1999social}, which is concerned with the impact of cognitive biases on social interaction. Another, more recent work by \citet{serfas2011cognitive} focused on the context of capital investment.  Closer to the domain of machine learning, in their article ``Psychology of Prediction'', \citet{kahneman1973psychology}  warned that cognitive biases can lead to violations of the Bayes theorem when people make fact-based predictions under uncertainty.  
These results directly relate to
inductively learned rules, since these are associated with measures such as confidence and support expressing the (un)certainty of the prediction they make.   
Despite some early work
\citep{michalski1969quasi,michalski1983theory}
showing the importance of study of cognitive phenomena for rule induction and machine learning in general, there has been a paucity of follow-up research.
In previous work \citep{furnkranz2019cognitive}, we have evaluated a selection of cognitive biases in the very specific context of whether minimizing the complexity or length of a rule will also lead to increased interpretability, which is often taken for granted in machine learning research.

In this paper, we attempt to systematically relate cognitive biases to the interpretation of machine learning results.
We anchor our discussion on inductively learned rules, but note in passing that a deeper understanding of human cognitive biases is important for all areas of combined human-machine decision making. We focus primarily on symbolic rules because they are generally considered to belong to the class of interpretable models, so that there is little general awareness that different ways of presenting or formulating them may have an important impact on the perceived trustworthiness, safety, or fairness of an AI system. In principle, our discussion also applies to rules that have been inferred by deduction, where, however, such concerns are maybe somewhat alleviated by the proved correctness of the resulting rules. 
To further our goal, we review twenty cognitive biases and judgmental heuristics whose misapplication can lead to biases that can distort the interpretation of inductively learned rules. The review is intended to help to answer questions such as: \emph{How do cognitive biases affect the human understanding of symbolic machine learning models?  What could help as a ``debiasing antidote''?}  

This paper is organized as follows.  Section~\ref{sec:background} provides a brief review of related work published at the intersection of rule learning and psychology.
Section~\ref{sec:example} motivates our study by showing an example of a learnt rule and discussing cognitive biases that can affect its plausibility. Section~\ref{sec:criteria} describes the criteria that we applied to select a subset of cognitive biases into our review, which eventually resulted in twenty biases. These biases and their respective effects and causes are covered in detail in Section~\ref{sec:review}. Section~\ref{sec:recommendations} provides a concise set of recommendations aimed at developers of rule learning algorithms and user interfaces.  In Section~\ref{sec:limitations} we state the limitations of our review and outline directions for future work. The conclusions summarize the contributions of the paper.

\section{Background and Related Work}
\label{sec:background}
We selected individual rules as learnt by many machine learning algorithms 
as the object of our study.  Focusing on simple artefacts---individual rules---as opposed to entire models such as rule sets or rule lists allows a deeper, more focused analysis since a rule is a small self-contained item of knowledge. Making a small change in one rule, such as adding a new condition, allows to test the effect of an individual factor. In this section, we first motivate our work by putting it into the context of prior research on related topics. Then, we proceed by a brief introduction to inductive rule learning (Section~\ref{ss:decrule}) and a brief recapitulation of previous work in cognitive science on the subject of decision rules (Section~\ref{ss:rules-cs}). Finally, we introduce cognitive biases (Section~\ref{ss:cbandib}) 
and  rule plausibility (Section~\ref{ss:measures}), which is a measure of rule comprehension.

\subsection{Motivation}
\label{sec:motivation}
In the following three paragraphs, we discuss our motivation for this review, and summarize why we think this work is relevant to the larger artificial intelligence community. 

\paragraph{Rules as Interpretable Models}
Given that neural networks and ensembles of decision trees are increasingly becoming the prevalent type of representation used in machine learning, it might be at first surprising that our review focuses almost exclusively on decision rules. The reason is that rules are widely used as a means for communicating explanations of a variety of machine learning approaches. 
In fact, quite some work has been devoted to 
explaining black-box models, such as neural networks, support vector machines and tree ensembles with interpretable surrogate models, such as rules and decision trees (for a survey on this line of work we refer, e.g., to \cite{guidotti2018survey}). 
As such a conversion typically also goes hand-in-hand with a corresponding reduction in the accuracy of the model, this approach has also been criticized \cite{StopExplaining}, and the interest in directly learning  rule-based models has recently renewed (see, e.g., \cite{jf:Book-Nada,wang2017bayesian,muggleton2018ultra,vojivr2018easyminer}).
\paragraph{Embedding cognitive biases to learning algorithms}
The applications of cognitive biases go beyond explaining existing machine learning models. For example, \citet{taniguchi2018machine} demonstrate how a cognitive bias can be embedded in a machine learning algorithm, achieving superior performance on small datasets compared to commonly used machine learning algorithms with ``generic'' inductive bias.

\paragraph{Paucity of research on cognitive biases in artificial intelligence}
Several recent position and review papers on explainability in Artificial Intelligence (xAI) recognize that cognitive biases play an important role in explainability research \citep{miller2018explanation,paez2019pragmatic}.
To our knowledge, the only systematic treatment of psychological phenomena applicable to machine learning is provided by the review of \citet{miller2018explanation}, which focuses on reasons and thought processes that people apply during explanation selection, such as causality, abnormality and the use of counterfactuals. This  authoritative review   observes that there are currently no studies that look at  cognitive biases in the context of  selecting explanations.
Because of the paucity of applicable research focusing on machine learning, the review of \citet{miller2018explanation} --- like the present paper --- takes the first step of applying influential psychological studies to explanation in the xAI context without accompanying experimental validation specific to machine learning.
While \citet{miller2018explanation} summarizes the main reasoning processes that drive generation and understanding of explanations, our review focuses specifically on cognitive biases as psychological phenomena that can distort the  interpretation of machine learning models if not properly accounted for. The role of bias mitigation in machine learning has been recently recognized in \cite{wang2019designing}, who describe four biases applicable to machine learning and for each propose a specific debiasing strategy. 
Our review is more comprehensive as we include twenty biases and we also provide a more detailed analysis of each of the biases included.

\subsection{Decision Rules in Machine Learning}
\label{ss:decrule}
An example of an inductively learned decision rule, which is a subject of the presented review,  is shown in Figure~\ref{fig:rule}.  

\begin{figure}[t]
\begin{Verbatim}[frame=single]
IF A AND B THEN C
   confidence=c and support=s

IF veil is white AND odor is foul THEN mushroom is poisonous
   confidence = 90%, support = 5%
\end{Verbatim}
\caption{Inductively learned rule}
\label{fig:rule}
\end{figure}

Following the terminology of \citet{jf:Book-Nada}, 
$A, B, C$ represent 
\emph{literals}, i.e., Boolean expressions which are composed of attribute name (e.g., \texttt{veil}) and its value (e.g., \texttt{white}). The conjunction of literals on the left side of the rule is called \emph{antecedent} or \emph{rule body}, the single literal predicted by the rule is called \emph{consequent} or \emph{rule head}.
Literals in the body are sometimes referred to as \emph{conditions} throughout the text, and the consequent as the \emph{target}.
While this rule definition  is restricted to conjunctive rules, other definitions, e.g., the formal definition given by \citet
{slowinski2006application}, also allow for negation and disjunction as connectives. 

Rules in the output of rule learning algorithms are most commonly characterized by two parameters, confidence and support.
The \emph{confidence} of a rule---sometimes also referred to as \emph{precision}---is defined as $\mathit{a}/(\mathit{a}+\mathit{b})$, where $\mathit{a}$ is the number objects that match both the conditions of the  rule as well as the consequent, and $\mathit{b}$ is the number of objects that match the antecedent but not  the consequent. 
The \emph{support} of a rule is either defined as $\mathit{a}/N$, where $N$ is the number of all objects (relative support), or simply as $a$ (absolute support). A related measure is \emph{coverage}, which is the total number of objects that satisfy the body of the rule ($a+b$).

The values of support and confidence  are often used as indications of how subjectively interesting the given rules is. Research has shown the utility of involving thresholds on a range of additional measures of significance  \cite{omiecinski2003alternative}. Out of the dozens of proposed formulas, the one most frequently adopted seems to be the \textit{lift} measure, which is a ratio of the confidence of the rule and the probability of occurrence of the head of the rule (not considering the body). If lift  is greater than 1, this indicates that the rule body and the rule head appear more often together than would correspond to chance.

In the special case of learning rules for the purpose of building a classifier, the consequent of a rule consists only of a single literal, the so-called \emph{class}. 
In this case, $a$ is also known as the number of \emph{true positives}, and $b$ as the number of \emph{false positives}.

Some rule learning frameworks, in particular association rule learning  \cite{APriori,AssociationRules-Book},
require the user to set thresholds for minimum confidence and support. 
Only rules with confidence and support values meeting or exceeding these thresholds are included on the output of rule learning and presented to the user.

Even though the terminology, ``support'' and ``confidence'', is peculiar to symbolic rule learning (in particular to association rule mining), the underlying concepts are universally adopted. For example, they are essentially equivalent to the terms ``recall'' and ``precision'' commonly used in information retrieval and correspond to the concepts of ``accuracy'' and ``coverage'' of general machine learning models.

\subsection{Decision Rules in Cognitive Science}
\label{ss:rules-cs}

Rules are used in commonly embraced models of human reasoning in cognitive science \citep{smith1992case,nisbett1993rules,pinker2015words}. They also closely relate  to Bayesian inference, which also frequently occurs in models of human reasoning. Consider the first rule of  Figure~\ref{fig:rule}.  
This rule can be interpreted as a hypothesis corresponding to the logical implication $A \land B \rightarrow C$. We can express  the plausibility of such a hypothesis in terms of Bayesian inference as the conditional probability $\Pr (C \mid A,B)$. This corresponds to the confidence of the rule, as used in machine learning and as defined above, 
and to the \emph{strength of evidence}, a term used by cognitive scientists \citep{Tversky27091974}.

Given that $\Pr(C \mid A,B)$ is a probability estimate computed on a sample, another relevant piece of information for determining the plausibility of the hypothesis is the robustness of this estimate. This corresponds to the number of instances for which the rule has been observed to be true. The size of the sample (typically expressed as a ratio) is known as rule support in machine learning and as the \emph{weight of the evidence} in cognitive science \citep{Tversky27091974}.\footnote{
	Interestingly, balancing the likelihood of the judgment and the weight of the evidence in the assessed likelihood  was already studied by \citet{keynes1922treatise} (according to \citet{Camerer1992}).}

Psychological research on hypothesis testing in rule discovery tasks has been performed in cognitive science at least since the 1960s. The seminal article by \citet{wason1960failure} introduced what is widely referred to as \emph{Wason's 2-4-6} task. Participants are given the sequence of numbers 2, 4 and 6 and asked to find out the rule that generated this sequence. In the search for the hypothesized rule, they provide the experimenter other sequences of numbers and the experimenter answers whether the provided sequence conforms to the rule, or not. While the target rule is simple ``ascending sequence'', people find it difficult to discover  this specific rule, presumably because they use the \emph{positive test strategy}, a strategy of testing a hypothesis by examining evidence confirming the hypothesis at hand rather than searching for disconfirming evidence \citep{klayman1987confirmation}. For example, if they have the hypothesis that the rule is a sequence of numbers increasing by two, they can provide a sequence 3-5-7, trying to confirm the hypothesis, rather than a sequence, such as 1-2-3, looking for an alternative hypothesis.

\subsection{Cognitive Bias}
\label{ss:cbandib}

According to the Encyclopedia of Human Behavior \citep{Wilke2012531}, the term cognitive bias was introduced in the 1970s by Amos Tversky and Daniel Kahneman \citep{Tversky27091974}, and is defined as a

\begin{quote}
 ``systematic error in judgment and decision-making common to all human beings which can be due to cognitive limitations, motivational factors, and/or adaptations to natural environments.''
\end{quote}

The narrow initial definition of cognitive bias as a shortcoming of human judgment was criticized by German psychologist  Gerd Gigerenzer, who started in the late 1990s the ``Fast and frugal heuristic'' program to emphasize ecological rationality (validity) of judgmental heuristics \cite{gigerenzer1999fastandfrugal}. According to this research program, cognitive biases often result from an application of a heuristic in an environment for which it is not suited rather than from problems with heuristics themselves, which work well in usual contexts.

In the present view, we define cognitive biases and associated phenomena  broadly. We include cognitive biases related to thinking, judgment, and memory. We also include descriptions of thinking strategies and judgmental heuristics that may result in cognitive biases, even if they are not necessarily biases themselves. 

\paragraph{Debiasing}
An important aspect related to the study of cognitive biases is the  validation of strategies for mitigating their effects in cases when they lead to incorrect judgment. A number of such \emph{debiasing} techniques have been developed, with researchers focusing intensely on the clinical and judicial domains (cf. e.g. \citep{lau2009can,croskerry2013cognitive,martire2014interpretation}), apparently due to costs associated with erroneous judgment in these domains. Nevertheless, general debiasing techniques can often be derived from such studies. 

The choice of an appropriate debiasing technique typically depends on the 
type of error induced by the bias, since 
this implies an appropriate debiasing strategy \citep{arkes1991costs}. \citet{larrick2004debiasing} recognizes the following three categories: psychophysically-based  error, association-based error, and strategy-based error.  The first two are attributable to the unconscious, automatic  processes, sometimes referred to as ``System~1''. The last one is attributed to reasoning processes (System~2) \cite{evans2013dual}.  
 For biases attributable to System~1, the most generic debiasing strategy  is to shift processing to the conscious System~2 \citep{lilienfeld2009giving}, \citep[p. 491]{shafir2013behavioral}. 

Another perspective on debiasing is provided by \citet{croskerry2013cognitive}, who organize debiasing techniques by their way of functioning, rather than the bias they address, into the following three categories: educational strategies, workplace strategies and forcing functions. While \citet{croskerry2013cognitive} focused on clinicians, our  review of debiasing aims to 
be used as a starting point for analogous guidelines for an audience of machine learning practitioners. For example, the general workplace strategies applicable in the machine learning context include group decision making, personal accountability, and planning time-out sessions to help slowing down. All of these strategies could lead to a higher probability of activating System 2 and thus reducing the biases which originate in the failure of System 1.

\paragraph{Function  and validity of cognitive biases}
\label{sec:functions}
The function of cognitive biases is a subject of scientific debate. 
According to the review of functional views by \citet{pohl2017cognitive}, there are three fundamental positions among researchers. The first group considers them as dysfunctional errors of the system, the second group as  faulty by-products of otherwise functional processes, and the third group as adaptive and thus functional responses.  According to \citet{pohl2017cognitive}, most researchers are in the second group, where cognitive biases are considered to be ``built-in errors of the human information-processing systems''.

In this work, we consider judgmental heuristics and cognitive biases as strategies that evolved to improve the fitness and chances of survival of the individual in particular situations or as consequences of such strategies. This defense of biases is succinctly expressed by  \citet{haselton2006paranoid}: ``Both the content and direction of biases can   be   predicted   theoretically   and   explained   by optimality when viewed through the long lens of evolutionary theory. Thus, the human mind shows good design, although it is designed for fitness maximization, not truth preservation.'' 

According to the same paper,  empirical evidence shows that cognitive biases are triggered or 
strengthened by environmental cues and context \citep{haselton2006paranoid}.
Given that the interpretation of machine learning results is a task unlike the simple automatic cognitive processes to which a human mind is adapted, cognitive biases are likely to have an influence upon it.

\subsection{Measures of Interpretability, Perceived and Objective Plausibility}
\label{ss:measures}
We claim that cognitive biases can affect the interpretation of rule-based models. However,  how does one measure interpretability? According to our literature review,  there is no generally accepted measure of interpretability of machine learning models. Model size, which was used in several studies, has recently been criticized \citep{freitas2014comprehensible,DBLP:conf/dis/StecherJF16,furnkranz2019cognitive} primarily on the grounds that the  model's syntactic  size does not capture any aspect of the model's  semantics. A particular problem related to semantics is the compliance to pre-existing expert knowledge, such as domain-specific monotonicity constraints.

In  prior work \cite{furnkranz2019cognitive}, we embrace the concept of \emph{plausibility} to measure interpretability. In the following, we will briefly introduce this concept because in the remainder of this article, we will use some material collected\footnote{\label{ftn:reuse}
In \cite{furnkranz2019cognitive}, we present quantitative results for several selected biases,  whereas the current article presents much broader review of the available literature. For illustrating some of the claims, we also make use of some of the 
textual responses from the participants, which were not featured in the above-mentioned work. 
} 
in user studies reported on in \cite{furnkranz2019cognitive} to illustrate some of the discussed biases and cognitive phenomena.
The word 'plausible' is defined according to the Oxford Dictionary of US English as ``seeming reasonable or probable'' and according to the Cambridge dictionary of UK English as ``seeming likely to be true, or able to be believed''.
We can link the 
inductively learned rule to the concept of ``hypothesis'' used in cognitive science. There is a body of work in cognitive science on analyzing the perceived plausibility of hypotheses \citep{gettys1978hypothesis,gettys1986plausibility,anderson2016analytical}.

In a recent review of interpretability definitions by \citet{bibal2016interpretability}, the term plausibility is not explicitly covered, but a closely related concept of \emph{justifiability} is stated to depend on  interpretability.   \citet{martens2011performance}  define justifiability as ``intuitively correct and in accordance with domain knowledge''.  By adopting plausibility,  we address the concern expressed in \citet{freitas2014comprehensible} regarding the need to reflect domain semantics when interpretability is measured.

\section{Motivational Example}
\label{sec:example}

\begin{figure}[H]
\includegraphics[width=12cm]{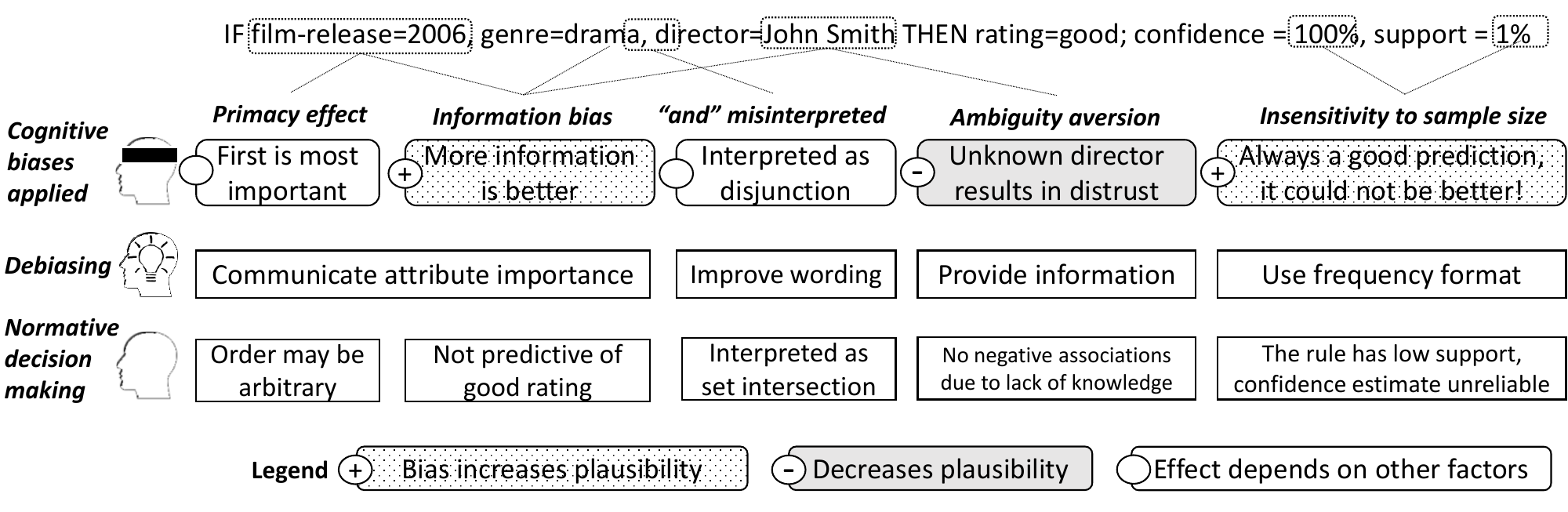}
\centering

\caption{Cognitive biases affecting perceived plausibility of example learnt rule} 

\label{fig:example}
\end{figure}

When an analyst\footnote{The prospective human users of rule models are often called ``analysts'' in this article.} evaluates the plausibility of a rule, a number of biases can be triggered by different facets of the rule. 
Consider, e.g., Figure~\ref{fig:example} which shows how the interpretation of a rule predicting whether a movie will get  a good rating based on its release date, genre, and director can be affected by various cognitive biases.
The analyst needs to evaluate whether each attribute and value is predictive for the target (good movie rating) and how large a set of movies it delimits. Additionally, the analyst needs to correctly process the syntactical elements in the rule (here AND and THEN), realizing that AND acts as a set intersection. Finally, the analyst needs to understand the confidence and support values and their trade-offs. 
Figure~\ref{fig:example} shows several illustrative cognitive biases for each of these processes. Two of them are discussed in greater detailed below, all of these (as well as many other) are covered in much greater detail in Section~\ref{sec:review}.

\paragraph{Information and unit bias}
According to  \emph{information bias}, more information  can make a rule look more plausible even if this information is irrelevant. 
In this case, the analyst may not know the director John Smith or any of his movies, but 
nevertheless, the rule that includes this condition may appear to be more plausible to the analyst than the same rule without this condition just because it involves more information. 
\paragraph{Insensitivity to sample size}
According to the \emph{insensitivity to sample size effect} \citep{Tversky27091974} there is a systematic bias in human thinking that makes humans overestimate the strength of evidence (confidence) and underestimate the weight of evidence (support).  
The example rule in Figure~\ref{fig:example} is associated with values of confidence and support that inform about the strength and weight of evidence. While it seems to have an excitingly high confidence (100\%), its low support indicates that this high value may be deceptive, as is sometimes the case for rule learning algorithms \citep{azevedo2007comparing}. 
Note that this also crucially depends on the absolute and not the relative support of this rule: if this rule originates from a database with a million of movies, a support of $1\%$ corresponds to 10,000 movies, whereas the same rule may only be based on a single movie in a database with 100 entries. Yet, the low support may be largely ignored by the analyst due to insensitivity to sample size.

\paragraph{Opposing effects of biases}
Sometimes the same piece of information can trigger opposing biases. Communicating the identity of the director of the movie can increase the plausibility due to the information bias, but if the specific director is not known to the user also decrease it through the ambiguity aversion bias. Figure~\ref{fig:example} also refers to the primacy effect  and misunderstanding of ``and'', where the impact on plausibility largely depends on the context. For  example, the year of the movie as the first provided information is overweighted due to the primacy effect, but  the overall  effect on plausibility will depend on how this information is perceived and aggregated by the analyst, which is determined by other factors.

\paragraph{Debiasing}
Whether the biases listed in Figure~\ref{fig:example}  apply depends, among other factors, on the analyst's background knowledge, quality of reasoning skills and statistical sophistication. 
An analysis of relevant literature from cognitive science not only reveals applicable biases but also sometimes provides methods for removing or limiting their effect (debiasing). Several debiasing techniques are also illustrated in Figure~\ref{fig:example}.
In principle, we found three categories of debiasing techniques:
\begin{inparaenum}
 \item Training users,
 \item Adapting learning algorithms,
 \item Adapting the representation of the model and/or the user interface. 
\end{inparaenum}

\paragraph{1. Training users} 
In \cite{fong1986effects} (cf. also \cite{nisbett1983use,reagan1989variations}) it was shown that training can significantly improve statistical
reasoning and help people better understand the importance of sample size ('law of large numbers'), which is instrumental for correctly interpreting statistical properties such as rule support and rule confidence.

\paragraph{2. Adapting learning algorithms} 
One possibility in terms of adaptation of learning algorithms is to compute confidence intervals for rule confidence
as proposed, e.g., in \cite{weiss2008statistical}. The support of a rule would then be---in a way---directly embedded into the presentation of rule confidence \cite{meo2010replacing}. Spurious rules with little statistical grounding 
may not be shown to the user at all.

\paragraph{3. Adapting the representation} 
A common way used in rule learning software for displaying rule confidence and support metrics is to use percentages, as in our example. Extensive research in psychology has shown that if frequencies are used instead, then the number of errors in judgment drops \citep{gigerenzer1996reasoning,gigerenzer1995improve}. Reflecting these suggestions, the hypothetical rule learnt from  our movies recommendation dataset could be presented as shown  in Figure~\ref{fig:supp-conf}.

\begin{figure}[t]
	\centering\fbox{%
		\begin{minipage}{0.95\textwidth}
			 \textbf{Mined rule:}
\begin{itemize}
			 \item[\ ]
 \texttt{IF film-release=2006 AND genre=drama} \\ \hspace*{6mm}\texttt{AND director="John Smith" THEN rating=good}\\ \texttt{\ \ \ confidence = 100\%, support = 1\%} 
			 \end{itemize}
			 
	   		\textbf{Verbalized rule:} 
	   		\begin{itemize}
	   		\item[\ ] If a film is released in \emph{2006} \textbf{and also} Genre is \emph{drama}  \textbf{and also} the director is John Smith then its rating is \emph{good}.
	   		\end{itemize}

   		\textbf{Improved explanation of rule:} 
   		
   		\begin{itemize}
   		    \item[\ ]
   		In our data, there are  \emph{2}  \emph{movies} which match the conditions of this rule. Out of these, \emph{2} are  correctly classified as being  \emph{good}. 
			The rule thus makes the correct prediction in  $\mathit{2/2=100\%}$  percent of cases, which corresponds to the confidence of the rule. The complete database contains \emph{200} \emph{movies}, out of these, the current rule correctly classifies~$2$. The support of the rule is thus $\mathit{2/200 =  1\%}$.
			\end{itemize}

		\end{minipage}
	}
		
	\caption{Suggested general frequency-based representation of an association rule 
	\label{fig:supp-conf}
	}
	\end{figure}
	
Rules can be presented in different ways (as shown), and depending on the way the information is presented, humans may perceive their plausibility differently.
In this particular example, confidence is no longer conveyed only as a percentage ``100\%'' but also using the expression ``2 out of 2''. Support is presented as an absolute number (2) rather than just a percentage (1\%).

A correct understanding of machine learning models can be difficult even for experts. 
In this section, we tried to demonstrate why addressing cognitive biases can play an important role in making the results of inductive rule learning more understandable. However, it should only serve as a motivational example, rather than a general guideline. In the remainder of this paper, the biases applied to our example will be revisited in greater depth, along with many other biases, and more concrete recommendations will be given.

\section{Scope of Survey}
\label{sec:criteria}
A number of cognitive biases  have been discovered, experimentally studied, and extensively described in the literature.
As \citet{pohl2017cognitive} states in a recent authoritative book on cognitive illusions: ``There is a plethora of phenomena showing that we deviate in our thinking, judgment and memory from some objective and arguably correct standard.'' 
This book covers 24 cognitive biases, and even 51 biases are covered by \citet{evans2007hypothetical}.

We first selected a subset of biases which would be reviewed. To select applicable biases, we looked for those that can interact with the following properties of rules, and their activation could result in an impact on perceived plausibility of rules:
\begin{inparaenum}
 \item rule length (the number of literals in an antecedent),
 \item rule interest measures (especially support and confidence),
 \item order of conditions in a rule and order of rules in the rule list, 
 \item specificity and predictive power of conditions (correlation with a target variable),
 \item use of additional logical connectives (conjunction, disjunction, negation),
 \item treatment of missing information (inclusion of conditions referring to missing values), and
 \item conflict between rules in the rule list.
\end{inparaenum}

Through a selection of appropriate learning heuristics, the rule learning algorithm can influence these properties. 
For example, most heuristics implement some form of a trade-off between the coverage or support of a rule, and its implication strength or confidence \cite{furnkranz2005roc,jf:Book-Nada}.

While doing the initial selection of cognitive biases to study, we tried to identify those most relevant for machine learning research matching our criteria. In the end, our review focused on a selection of 20 judgmental heuristics and cognitive biases. Future work might focus on expanding the review with additional relevant biases, such as labelling and overshadowing effects \citep{pohl2017cognitive}.

\section{Review of Cognitive Biases}
\label{sec:review}
\begin{figure}[t]
\includegraphics[width=12cm]{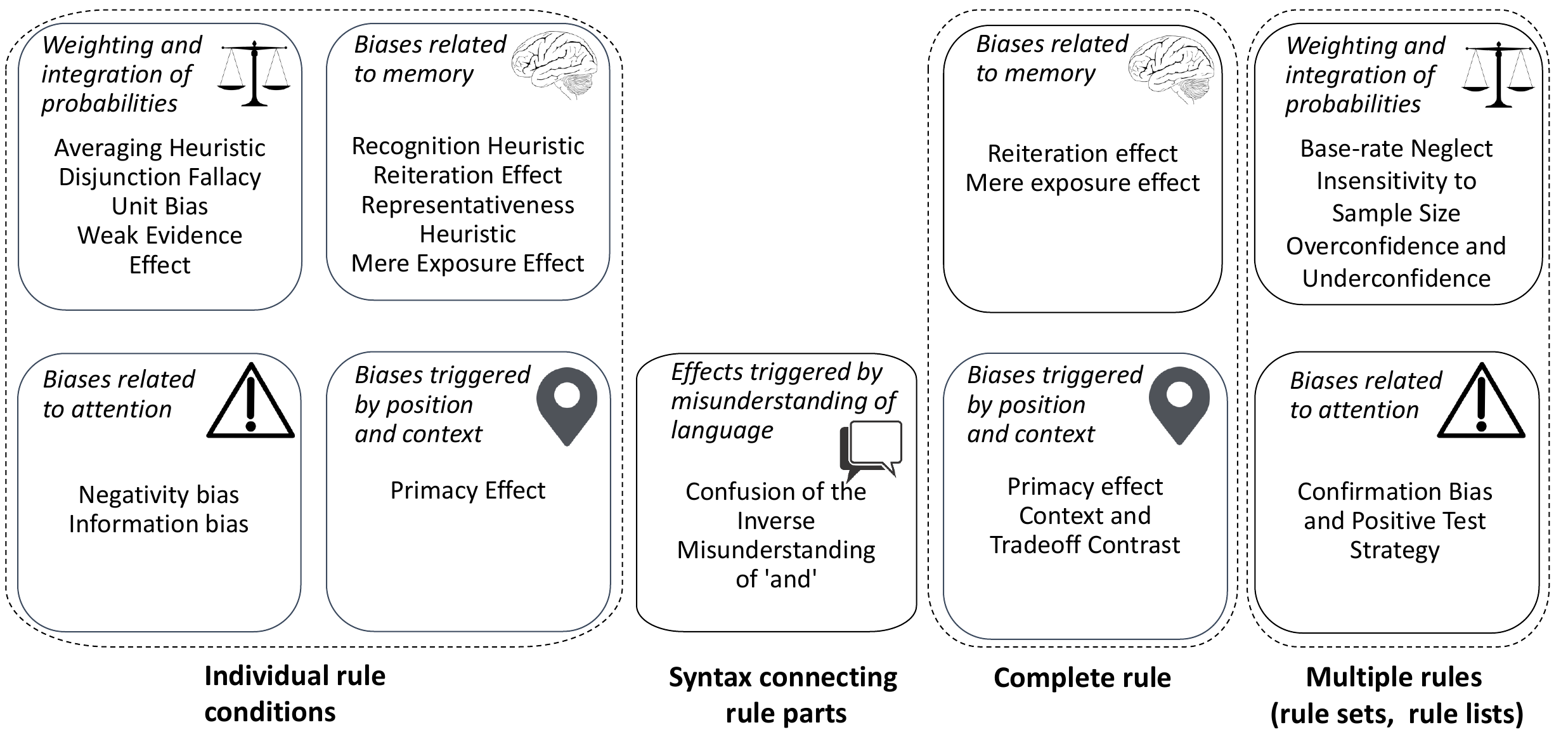}
\centering

\caption{Cognitive biases grouped by the affected or triggering rule model element and by their underlying mechanism} 

\label{fig:overview}
\end{figure}

In this section, we cover a selection of twenty cognitive biases.
For all of them, we include a short description including an example of a study demonstrating the bias and its proposed explanation.
We pay particular attention to their potential effect on the interpretability of rule learning results, which has not been covered in previous works. Figure~\ref{fig:overview} shows a high-level overview of the results of our analysis. The figure organizes the surveyed biases according to the primary affected element of rule models, ranging from conditions (literals) as the basic building block, to entire rules and rule models. The second perspective conveyed in the figure relates to the underlying mechanism of the biases.

In a recent scientometric survey of research on cognitive biases in information systems \citep{DBLP:conf/ecis/FleischmannABH14}, no articles are mentioned that aim at machine learning.  For general information systems research, the authors claim that ``most articles' research goal [is] to provide  an  explanation  of  the  cognitive  bias  phenomenon  rather  than to  develop  ways  and  strategies  for  its  avoidance  or targeted  use''. In contrast, our review aims at the advancement of the field beyond the explanation of applicable phenomena, by also discussing specific debiasing techniques.

For all cognitive biases, we thus suggest a debiasing technique that could be effective in aligning the perceived plausibility of the rule with its objective plausibility. While we include a description only of the most prominent debiasing strategies for each bias, it is possible that some of the debiasing strategies may be more general and could be effective for multiple biases.   

The utility of this article for the reader would increase if concrete proposals for debiasing machine learning (rule learning) results were included. However,  there is a paucity of applicable work on debiasing techniques applied specifically to machine learning (or directly on rule learning), and the invention of  a rule-learning-specific debiasing technique for each of the twenty surveyed biases is out of the scope of  this article. We, therefore, decided to introduce only one debiasing technique developed specifically for rule learning, choosing an approach that is to  our knowledge the most well studied. 

We chose the ``frequency format'' debiasing method, which has been documented to reduce the number of bias-induced judgment errors across a variety of different tasks as supported by a body of psychological studies. We adapted this method for rule learning and used in a user study reported in \cite{furnkranz2019cognitive}. Since then, it has been subject of at least one other user study focused specifically on debiasing methods for rule learning. This method is introduced already in Section~\ref{ss:reprh} in the context of the motivating Linda problem and the representativeness heuristic. However, since it provides a way for  expressing rule confidence and rule support, it is primarily intended as a debiasing method for the base rate neglect (Section~\ref{ss:baseratefallacy}) and the insensitivity to sample size (Section~\ref{ss:ins}). The proposed generalized adaptation to rule learning also adopts recommendations from psychological research for addressing the misunderstanding of \emph{'and'} (Section~\ref{ss:misunderstandingAnd}).

An overview of the main features of the reviewed cognitive biases is presented in Table~\ref{tbl:bias-overview}. Note that the  debiasing techniques that we describe have only limited grounding in applied psychological research and require further validation, since as \citet{lilienfeld2009giving} observe, there is a general paucity of research on debiasing in psychological literature, and the existing techniques suffer from a lack of theoretical coherence and mixed research evidence concerning their efficacy.

\begin{sidewaystable}
\centering
\begin{tabular}{p{8cm}p{7cm}p{6cm}} \toprule
phenomenon                                                                                                                                                                                                                                                               & implications for rule-learning                                                                                                                                                                                                                                                                               & debiasing technique                                                                                                                               \\ \midrule
\emph{Conjunction Fallacy and Representativeness Heuristic} (Section~\ref{ss:reprh}): A judgmental heuristic using prototypicality for judgment of probability, which may lead people to judge a conjunction of two statements to be more probable than one of the two statements. & Overestimate the probability of condition representative of rule consequent                                                                                                                                                                                                                                  & Use natural frequencies instead of ratios or probabilities, engage System 2, provide training in logic                                            \\  \hline
\emph{Misunderstanding of 'and'} (Sec.~\ref{ss:misunderstandingAnd}): The misunderstanding of conjunction 'and' used to denote an intersection as indicating a collection of sets or generally as some other than the intended meaning.                                                                                       & People interpret 'and' differently than logical conjunction                                                                                                                                                                                                                                                  & present conditions as propositions rather than categories                                                                                         \\ \hline
\emph{Averaging Heuristic}: A judgmental heuristic using the average of probabilities of two events for estimation of the probability of a conjunction of the two events.                                                                                                       & Probability of antecedent as the average of probabilities of conditions                                                                                                                                                                                                                                      & Reminder of probability theory                                                                                                                    \\ \hline
\emph{Disjunction Fallacy}: Judging the probability of an event as higher than the probability of a union of the event with another event.                                                                                                                                      & Prefer more specific conditions over less specific                                                                                                                                                                                                                                                           & Inform on taxonomical relation between conditions; explain benefits of higher support                                                             \\ \hline
\emph{Base-rate Neglect}: The tendency to underweight evidence provided by base rates.                                                                                                                                                                                        & Neglect prior probability of the head of the rule                                                                                                                                                                                                                                                            & Show and explain the value of lift measure                                                                                                        \\ \hline
\emph{Insensitivity to Sample Size}: The tendency to underestimate the benefits of larger samples.                                                                                                                                                                              & Analyst does not realize the increased reliability of confidence estimate with increasing value of support                                                                                                                                                                                                   & Present support as absolute number; show confidence (reliability) intervals for the value of confidence  \\ \hline
\emph{Confirmation Bias and Positive Test Strategy}: The tendency to seek supporting evidence for one's current hypothesis.                                                                                                                                                     & Rules confirming analyst's prior hypothesis are \`{}\`{}cherry picked''                                                                                                                                                                                                                                      & Guidance to consider evidence for and against hypothesis; education about the bias; interfaces making users slow down                    \\ \hline
\emph{Availability heuristic}: Perceived frequency of a class is determined by the ease with which its instances come to mind & Rules  for which instances are easily recalled are considered as more plausible & Explain why for some combinations of conditions the instances are easily recalled  and not for others
 \\ \hline
\emph{Reiteration Effect}: The increase of perceived believability following repetition.                                                                                                                                                                                        & Presentation of redundant rules or conditions increases plausibility                                                                                                                                                                                                                                         & rule pruning; clustering; explaining overlap                                                                                                      \\ \hline
\emph{Mere Exposure Effect}: The increase of liking following repetition.                                                                                                                                                                                                       & Repeated exposure (even subconscious) results in increased preference                                                                                                                                                                                                                                        & Changes to user interfaces that limit subliminal presentation of rules                                                                            \\ \hline
\emph{Overconfidence and Underconfidence}: The tendency to be overconfident (or underconfident) in one's judgments.                                                                                                                                                             & Rules with small support and high confidence are \`{}\`{}overrated''                                                                                                                                                                                                                                         & Present less information when not relevant with rule pruning algorithms, feature selection, limiting rule length; actively present conflicting rules/knowledge.    \\ \hline
\emph{Recognition Heuristic}: A judgmental heuristic using recognition for judgment of frequency, size, or other attributes.                                                                                                                                                    & Recognition of attribute or its value increases preference                                                                                                                                                                                                                                                   & More time; knowledge of attribute/value                                                                                                           \\ \hline
\emph{Information Bias}: The tendency to seek information even when it is not relevant or helpful.                                                                                                                                                                              & belief that more information (rules, conditions) will improve decision making even if it is irrelevant                                                                                                                                                                                                       & Communicate attribute importance                                                                                                                  \\ \hline
\emph{Ambiguity Aversion}: The tendency to prefer known risks over unknown risks.                                                                                                                                                                                                & Prefer rules without unknown conditions                                                                                                                                                                                                                                                                      & Increase user motivation; instruct users to provide textual justifications                                                                        \\ \hline
\emph{Confusion of the Inverse}: The mistake of confusing the confidence of an implication A $\rightarrow$ B with its inverse B $\rightarrow$ A.                                                                                                                                                        & Confusing the difference between the confidence of the rule $\Pr(\textrm{consequent} \mid \textrm{antecedent})$  with $\Pr(\textrm{antecedent} \mid \textrm{consequent})$  & Training in probability theory; unambiguous wording                                                                                               \\ \hline
\emph{Context and Tradeoff Contrast}: An effect of the context of a choice on preferences of available options.                                                                                                                                                                 & Preference for a rule is influenced by other rules                                                                                                                                                                                                                                                           & Removal of rules, especially of those that are strong, yet irrelevant                                                                             \\ \hline
\emph{Negativity Bias}: The tendency to overweight negative information over positive information of the same strength.                                                                                                                                                                          & Words with negative valence in the rule make it appear more important                                                                                                                                                                                                                                        & Review words with negative valence in data, and possibly replace with neutral alternatives                                                        \\ \hline
\emph{Primacy Effect}: A disproportionate effect of initial information on the final assessment.                                                                                                                                                                                & Information (rules, conditions) presented first has the highest impact                                                                                                                                                                                                                                                           & Education on the bias; resorting; rule annotation                                                                                                 \\ \hline
\emph{Weak Evidence Effect}: Weak argument in favor of a statement can lead to decreased believability of the statement.                                                                                                                                                           & Condition only weakly perceived as predictive of target decreases plausibility                                                                                                                                                                                                                               & Numerical expression of strength of evidence; omission of weak predictors (conditions)                                                            \\ \hline
\emph{Unit Bias}: The tendency to give a similar weight to each unit rather than weigh it according to its size.                                                                                                                                                                & Conditions are perceived to have same importance                                                                                                                                                                                                                                                             & Inform on discriminatory power of conditions                                                                                                      \\ \bottomrule
\end{tabular}
\caption{Summary of analysis of cognitive biases}
\label{tbl:bias-overview}
\end{sidewaystable}
\subsection{Conjunction Fallacy and Representativeness Heuristic}
\label{ss:reprh}

The conjunction fallacy refers to a judgment that is inconsistent with \emph{the conjunction rule} -- the probability of conjunction, $\Pr(A, B)$, cannot exceed the probability of its constituents, $\Pr(A)$ and $\Pr(B)$. It is often illustrated with the ``Linda'' problem in the literature \citep
{tversky1983extensional}. 
In the Linda problem, depicted in Figure~\ref{fig:Linda}, subjects are asked to 
compare conditional probabilities   $\Pr(B \mid L)$ and $\Pr(F,B \mid L)$, where $B$ refers to ``bank teller'', $F$ to  ``active in feminist movement'' and $L$ to the description of Linda \citep{bar1991commentary}.  

\begin{figure}[ht!]
\begin{Verbatim}[frame=single]
  Linda is 31 years old, single, outspoken, and very bright.
  She majored in philosophy. As a student, she was deeply 
  concerned with issues of discrimination and social justice, 
  and also participated in anti-nuclear demonstrations.

  Which is more probable?

  (a) Linda is a bank teller.
  (b) Linda is a bank teller and is active in the
      feminist movement.
 
\end{Verbatim}      

\caption{Linda problem}
\label{fig:Linda}
\end{figure}

Multiple studies have shown that people tend to consistently select the second  hypothesis as more probable, which is in  conflict with the conjunction rule. In other words, it always holds for the Linda problem that  $$\Pr(F,B \mid L) \leq \Pr(B \mid L).$$ Preference for the alternative $F \wedge B$ (option (b) in Figure~\ref{fig:Linda}) is thus always a logical fallacy. For example, \citet{tversky1983extensional}  report that 85\% of their subjects indicated (b) as the more probable option for the Linda problem. The conjunction fallacy has been shown across multiple settings (hypothetical scenarios, real-life domains), as well as for various kinds of subjects (university students, children, experts, as well as statistically sophisticated individuals) \citep{tentori2012conjunction}. 

The conjunction fallacy is often 
explained by the use of the representativeness heuristic \citep{kahneman1972subjective}. The representativeness heuristic refers to the tendency to make judgments based on similarity,  based on the rule ``like goes with like'', which is typically used to determine whether an object belongs to a specific category. When people use the representativeness heuristic, \emph{``probabilities are evaluated by the degree to which A is representative of B, that is by the degree to which A resembles B''} \citep{Tversky27091974}.
This heuristic provides people with means for assessing the probability of an uncertain event. It is used to answer questions such as ``What is the probability that object A belongs to class B? What is the probability that event A originates from process B?'' \citep{Tversky27091974}.

The representativeness heuristic is not the only explanation for the results of the conjunction fallacy experiments. \citet{hertwig2008conjunction} hypothesized that the fallacy is caused by ``a  misunderstanding about conjunction'', in other words by a different interpretation of ``probability'' and ``and''  by the subjects than assumed by the experimenters. The validity of this alternative hypothesis has been subject to criticism  \citep{tentori2012conjunction}, nevertheless,  some empirical evidence suggests that the problem of the correct understanding of ``and'' is of particular importance to rule learning \citep{furnkranz2019cognitive}.

Recent research has provided several explanations for conjunctive and disjunctive  (cf.\ Section~\ref{ss:disjfal})  fallacies, such as Configural Weighting and Adding (CWA)+ theory \citep{nilsson2009linda}, applying principles of quantum cognition \citep{bruza2015quantum} and inductive confirmation theory \citep{tentori2013determinants}. In the following, we will focus on the CWA theory. 
CWA essentially assumes that the causes of conjunctive and disjunctive fallacies relate to the fact that decision makers perform  a weighted average instead of multiplication of the component probabilities. For conjunctions, weights are set so that  more weight is assigned to the lower component probability.  For  disjunctive probabilities,
more weight is assigned to the likely component. This assumption was  verified in at
least one study \citep{fisk2002judgments}.  For more discussion of the related averaging heuristic, cf. Section~\ref{ss:averaging}.

There is also a body of research that accounts the results of the Linda experiment to the violation of conversational maxims. \citet{politzer1991conjunction} argue that since the two options given to the participants are nested, it compels the participants to interpret the first choice (A) as A and not-B. According to this interpretation, subjects' typical responses are in line with logical principles.
\citet{donovan1997difficulty} have conducted follow-up experiments presenting completely disclosing information and accounting for whether a particular choice was made for a good (logical reasoning) or wrong reason (representativeness heuristic). The results have shown that ``conjunction errors'' cannot be to large extent attributed to the violation of the implicit conversational norms.

\paragraph{Implications for rule learning}
Rules are not composed only of conditions, but also of an outcome (the value of a target variable in the consequent). A higher number of conditions generally allows the rule to filter a purer set of objects with respect to the value of the target variable than a smaller number of conditions.
Application of representativeness heuristic can affect the human perception of rule plausibility in that rules that contain conditions (literals) more ``representative'' of the user's mental image of the concept may be preferred even in cases when their objective discriminatory power may be lower.

In the example depicted in Figure~\ref{fig:reprEx}, we show a hypothesized effect of the representativeness heuristic.
The example shows several justifications of user choices collected in the experiments reported in \cite{furnkranz2019cognitive}.
As follows from the second response, for a rule to be found plausible, it needs not only to   contain isolated representative literals but the collection of literals as a whole needs to provide a coherent impression of a representative description.

\begin{figure}[t]
\begin{center}
\fbox{\begin{minipage}{12cm}
Users recruited via crowdsourcing were asked to say which of the  following two rules is more plausible, to what degree, and why:

\begin{itemize}
    \item \textbf{Rule 1:} if the mushroom has the following properties (simultaneously): 
    gill size is \emph{broad} \textbf{and}
    stalk surface above ring is  \emph{smooth} \textbf{and}
    mushroom has  \emph{one ring}
  \textbf{and} spore color is \emph{brown}  then the mushroom is edible
\item \textbf{Rule 2:}  odour is  \emph{anise} then the mushroom is edible
\end{itemize}
Similarly, in another task, we asked users to assess a shorter rule with just one condition (creosote odour) and a longer rule, both predicting the poisonous class. 
In both tasks, the justifications provided by multiple participants referred to smell, which is representative of the mental image of edibility:

\begin{itemize}
    \item \emph{``Anise smells nice and usually nice smelling things aren't poisonous.''}
    \item ``\emph{odor alone is not enough properties to make a determination.}''
    \item ``\emph{An unpleasant odour is the red flag for this statement being more likely.}'' 
    \item ``\emph{The smell of the mushroom is poisonous smell.}''
\end{itemize}

\emph{Example adapted from data gathered during a user study partly reported in \cite{furnkranz2019cognitive}.}
\end{minipage}}
\end{center}
\caption{Example of the hypothesized effect of the representativeness heuristic.}
\label{fig:reprEx}
\end{figure}

\paragraph{Debiasing techniques}

A number of factors that decrease the proportion of subjects exhibiting the conjunction fallacy have been identified:
\citet{charness2010conjunction} found that the number of participants committing the fallacy is reduced under a monetary incentive. Such an addition was reported to drop the fallacy rate in their study from 58\% to 33\%. The observed rate under a monetary incentive suggests smaller importance of this problem for important real-life decisions. \citet{zizzo2000violation} found that unless the decision problem is simplified, neither monetary incentives nor feedback can ameliorate the fallacy rate. A reduced task complexity is a precondition for monetary incentives and feedback to be effective. 

\citet{stolarz1996conjunction} observed that the rate of fallacies is reduced but still strongly present when the subjects receive training in logic.   \citet{gigerenzer1996reasoning}, as well as \citet{gigerenzer1995improve}, showed that the rate of fallacies can be reduced or even eliminated by presenting the problems in terms  of frequencies rather than probabilities.
To illustrate this debiasing measure, Figure~\ref{fig:LindaFrequency} presents  a reformulation of the Linda problem in the frequency format, and Figure~\ref{fig:supp-conf} presents a reformulation of a discovered rule in the frequency format with wording, which is based on (but does not exactly correspond to) a formulation used in experiments reported in \cite{furnkranz2019cognitive}. 

Note that the original proposal of frequency formats \cite{gigerenzer1995improve} suggests \emph{replacing} ratios with frequencies, however, in Figure~\ref{fig:supp-conf}, we provide frequencies \emph{alongside} the precomputed ratios of confidence and support. 
The reason is that it has been empirically shown that using only the frequency format (based on actual values from the rule's confusion matrix as in \cite{pagliaro2020cognitive}) is not helpful to users. In experiments on the UCI soybean dataset reported in \cite[p. 51]{pagliaro2020cognitive},  the users were shown the following explanations for values of confidence and support: \emph{``In our data, which contains 631 observations, 71 of these match the conditions of this rule. Out of these, 55 are predicted as having the phytophthora root disease''}. The results of the user study showed that this led to a higher number of errors than if percentage values of  confidence and support were shown, which is interpreted in \cite{pagliaro2020cognitive} as a result of the mental computation of the ratios by the participants. It is also not an option to use the value 100 or 1000 as a reference value as recommended in \cite{gigerenzer1995improve} and demonstrated in Figure~\ref{fig:LindaFrequency}. The reason is that while this formulation is useful for expressing confidence, it  distorts the user's understanding of support. As a result of this analysis, we arrived at the suggested wording of the debiasing presented in  Figure~\ref{fig:supp-conf}. We emphasize that this is only a proposal, which needs further experimental validation with, as we envisage, a between-subject methodology.

\begin{figure}[ht!]
\begin{Verbatim}[frame=single]
Linda is 31 years old, single, outspoken, and very bright. She
majored in philosophy. As a student, she was deeply concerned
with issues of discrimination and social justice, and also
participated in anti-nuclear demonstrations. Of 100 people
like Linda, how many are ...? 
\end{Verbatim}    
\caption{Linda problem -  frequency reformulation in \citep{mellers2001frequency}}
\label{fig:LindaFrequency}
\end{figure}

\citet{nilsson2009linda} present a computer simulation showing that when the component probabilities are not precisely known, averaging often provides an equally good alternative to the normative computation of probabilities (cf. also \citet{juslin2009probability}). The reason is that the normatively correct multiplicative integration can amplify the errors  more than linear additive integration. This computational model could be possibly adapted to detect a high risk of fallacy,  corresponding to the case when the deviation between the perceived probability and the normative probability is high.

In the context of explaining medical machine learning results, \citet{wang2019designing}  propose to show prototypes of patient instances for each diagnosis. In rule learning, this would imply  extending the user interface with a functionality that would show instances in data supporting the rule as well as the false positives -- instances matching the antecedent, but not the rule consequent.

\subsection{Misunderstanding of ``and''}
\label{ss:misunderstandingAnd}
The misunderstanding of ``and'' refers to a phenomenon affecting the syntactic comprehensibility of the logical connective ``and''. As discussed by \citet{hertwig2008conjunction}, ``and'' in natural language can express several relationships, including temporal order, causal relationship, and most importantly, can also indicate a collection of sets\footnote{As in ``He invited friends and colleagues to the party''} as well as their intersection. People can therefore interpret ``and'' in a different meaning than intended.

For example, according to the two experiments reported by \citet{hertwig2008conjunction},  the conjunction ``bank teller and active in the feminist  movement'' used in the Linda problem (cf.\ Section~\ref{ss:reprh}) was found by about half of subjects as ambiguous---they explicitly asked the experimenter how ``and'' was to be understood. Furthermore, when participants indicated how they understood ``and'' by shading Venn diagrams, it turned out that about a quarter of them interpreted ``and'' as union rather than an intersection, which is usually assumed by experimenters using the Linda problem.

\paragraph{Implications for rule learning}
The formation of conjunctions via ``and'' is a basic building block of rules. Its correct understanding is thus important for effective communication of results of rule learning. 
Existing studies suggest that the most common type of error is understanding ``and'' as a union rather than intersection. In such a case, a rule containing multiple ``ands'' will be perceived as having a higher support than it actually has. Each additional condition will be incorrectly perceived as increasing the coverage of the rule. This implies higher perceived plausibility of the rule.

Similarly as in the previous section (Figure~\ref{fig:reprEx}), we can illustrate the occurrence of this bias on user feedback collected as part of experiments conducted in a research project, partly reported in \cite{furnkranz2019cognitive}.
There, example responses that support the misunderstanding of 'and' include: \emph{``Rule 1 contains twice as many properties as rule 2 does for determining the edibility of a mushroom so that makes it statistically twice as plausible, hence much higher probability of being believable''}, \emph{``An extra group increases the likelihood.''}
Misunderstanding of ``and'' will thus generally increase the preference of rules with more conditions.

\paragraph{Debiasing techniques}
According to \citet{sides2002reality} ``and'' ceases to be ambiguous when it is used to connect propositions  rather than categories. The authors give the following example of a sentence which is not prone to misunderstanding: ``IBM stock will rise tomorrow and Disney stock will fall tomorrow.''
A similar wording of rule learning results, as depicted in Figure~\ref{fig:supp-conf} (page~\pageref{fig:supp-conf}), may be preferred, despite its verbosity. In the figure, conditions are represented as propositions (``film release is \emph{2006} \textbf{and also} $\ldots$'') as opposed to categories. These are otherwise often used in machine learning due to the use of various feature recoding methods, which would result in less understandable  ``film\_release \textbf{and} $\ldots$''.

\citet{mellers2001frequency} showed that using ``bank tellers who are feminists'' or ``feminist bank tellers'' rather than ``bank tellers and feminists'' as a category in the Linda problem (Figure~\ref{fig:Linda}) might reduce the likelihood of committing the conjunction fallacy.
It follows that using different wording 
such as ``and also'' might also help reduce the danger of a misunderstanding of ``and''.

Representations that visually express the semantics of ``and'' such as decision trees may be preferred over rules, which do not provide such visual guidance.\footnote{We find limited grounding for this proposition in the following: Conditions connected with an arch in a tree are to be interpreted as simultaneously valid (i.e., arch means conjunction).  A recent empirical study on the comprehensibility of decision trees \citep{Piltaver2016333} does not consider the ambiguity of this notation to be a systematic problem among the surveyed users.}

\subsection{Averaging Heuristic}
\label{ss:averaging}
While the conjunction fallacy is most commonly explained by the operation of the representativeness heuristic, the averaging heuristic provides an alternative explanation: it suggests that people evaluate the probability of a conjuncted event as the average of probabilities of the component events \citep{fantino1997conjunction}. 
As reported by \citet{fantino1997conjunction}, in their experiment ``approximately 49\% of variance in subjects' conjunctions could be accounted for by a model that simply averaged the separate component likelihoods that constituted a particular conjunction.''

\paragraph{Implications for rule learning}
When applying the averaging heuristic, an analyst may not fully realize the consequences of the presence of a low-probability condition for the overall likelihood of the set of conditions in the antecedent of the rule. 

Consider the following example: Let us assume that the learning algorithm only adds independent conditions that have a probability of $0.8$, and we compare a 3-condition rule to a 2-condition rule. 
Averaging would evaluate  both rules equally, because both have an average probability of $0.8$. Correct computation of the joint probability, however, shows that the longer rule is considerably less likely ($0.8^3$ vs.\ $0.8^2$ because all conditions are assumed to be independent).
We expect that the averaging heuristic could be triggered both if the probability of the conditions is explicitly shown and if it is only based on internal knowledge of the decision maker. 

Averaging can also affect same-length rules.
Continuing our example, if we compare the above 2-condition rule with another rule with two features with more diverse probability values, e.g., one condition has $1.0$ and the other has $0.6$, then averaging would again evaluate both rules the same, but in fact the correct interpretation would be that the rule with equal probabilities is more likely than
the other ($0.8^2 > 1.0 \times 0.6$). 
In this case, the low 0.6 probability in the new rule would ``knock down'' the normative conjoint probability below the one of the rule with two 0.8 conditions.

\paragraph{Debiasing techniques}
Experiments conducted by \citet{zizzo2000violation} showed that prior knowledge of probability theory, and a direct reminder of how probabilities are combined,  are effective tools for decreasing the incidence of the conjunction fallacy, which is the hypothesized consequence of the averaging heuristic.
A specific countermeasure for the biases caused by linear additive integration (weighted averaging) is the  use of logarithm formats. Experiments conducted by \citet{juslin2011reducing} show that recasting probability computation in terms of logarithm formats, thus requiring additive rather than multiplicative integration, improves probabilistic reasoning.
However, it should be noted that the use of the  logarithm format may require additional training and also may lead to increased mental effort for the user. As a result, this is an example of a debiasing technique where the associated costs may outweigh the benefits.

\subsection{Disjunction Fallacy }
\label{ss:disjfal}
The disjunction fallacy refers to a judgment that is inconsistent with the disjunction rule, which states that the probability $\Pr(X)$ cannot be higher than the probability  $\Pr(Z)$, where $Z$ is a union of event $X$ with another event $Y$ (i.e., $X\cup Y$). 

In experiments reported by \citet{bar1993alike}, $X$ and $Z$ were nested pairs of categories, such as Switzerland and Europe.  Subjects read descriptions of people such as ``Writes letter home describing a country with snowy wild mountains, clean streets, and  flower decked porches. Where was the letter written?''  It follows that since Europe contains Switzerland, Europe must be more likely than Switzerland. However, Switzerland was rated as the more likely place by about 79\% of the participants \citep{bar1993alike}. 

The disjunction fallacy is considered as another consequence of the representativeness heuristic \citep{bar1993alike}: ``Which of
two events---even nested events---will seem more probable is better predicted by their representativeness than by their scope, or by the  level in the category hierarchy in which they are located.''
The description in the example is more representative of Switzerland than of Europe, so when people use representativeness as the basis for their judgment, they judge Switzerland to be a more likely answer than Europe, even though this judgment breaks the disjunction rule.

\paragraph{Implications for rule learning}
In the context of data mining, it can be the case that the feature space is hierarchically ordered. The analyst can thus be confronted with rules containing attributes (literals) on multiple levels of granularity. Due to the disjunction fallacy, the analyst will generally prefer rules containing more specific attributes, which can result in a preference for rules with fewer backing instances and thus in weaker statistical validity.

Similarly as in  some of the previous sections, we can illustrate the occurrence of this bias on user feedback collected as part of experiments conducted in \cite{furnkranz2019cognitive}.
There, multiple responses  may be possibly linked to the disjunction fallacy (preference for specific attributes).  In the absence of other deciding factors, one group of subjects preferred longer rules on the grounds that these are considered as more ``descriptive'', ``specific'', ``less broad'' or having ``more data''.

\emph{``There's nothing that stands out as an ``obvious'' indicator of toxicity, so I've gone for a weak preference for Rule 2 as it's describing a smaller number of species than Rule 1 and thus likely to be the more accurate of the two''}, \emph{``An extra group increases the likelihood.''}

\paragraph{Debiasing techniques}
When asked to assign categories to concepts (such as a place of origin of a letter) under conditions of certainty, people are known to prefer a specific category to  a more general category that subsumes it, but only if the specific category is considered representative  \citep{bar1993alike}: ``whenever an ordering of events by  representativeness  differs from their ordering  by set inclusion, there is a  potential for an extension fallacy to occur.''
From this observation a possible debiasing strategy emerges: making the analysts aware of the taxonomical relation of the individual attributes and their values. For example, the user interface can  work with the information that Europe contains Switzerland, possibly actively notifying the analyst on the risk of falling for the disjunctive fallacy. This intervention can be complemented by ``training in rules'' \citep{larrick2004debiasing}. 
In this case, the analysts should be explained the benefits of a larger supporting sample  associated with more general attributes. 

\subsection{Base-rate Neglect}
\label{ss:baseratefallacy}
People tend to underweight the evidence provided by base rates, which results in the so-called \emph{base-rate neglect}. For example, \citet{kahneman1973psychology} gave participants a description of a person who was selected randomly from a group and asked them whether the person is an engineer or a lawyer. Participants based their judgment mostly on the description of the person and paid little consideration to the occupational composition of the group, 
even though the composition was provided as part of the task and should play a significant role in the judgment.

\citet{kahneman1973psychology} view the base-rate neglect as a possible consequence of the representativeness heuristic \citep{kahneman1972subjective}. When people base their judgment of an occupation of a person mostly on the similarity of the person to a prototypical member of the occupation, they ignore other relevant information such as base rates, which results in the base-rate neglect.

\paragraph{Implications for rule learning}
Considering an inductively learnt rule of the form 
\texttt{IF} $A$ \texttt{AND} $B$ \texttt{THEN} $C$, the concept of base rate corresponds to the prior probability of the head of the rule $\Pr(C)$. While this information is important for assessing the plausibility of the rule, the two most commonly used statistical measures of rule quality, support and confidence, are not sufficient to communicate the base rate to the  user. 

Let's consider the following two rules generated from a hypothetical dataset containing personal characteristics of 1,000 people along with their occupation, which is the target variable. 
Out of these, there are 18 engineers and 72 lawyers, and the rest are different professions, which determines the respective base rates of $1.8$\% for engineers and $7.2$\% for lawyers.

Let us now consider the following two rules, both  having 1.8\% support. Since the total number of instances is 1.000, each of these two rules  correctly classifies $0.018 \times 1000=18$ instances in the training data. The confidence of both rules is 90\%, which means that each rule misclassifies 10\% of instances covered by its antecedent.
\mybox{
\noindent\textbf{Example.} 
\begin{itemize}
 \item[R1] \texttt{IF writing is average AND intelligence is high AND eloquence is average THEN profession is engineer \\
 confidence =  90\%, support = 1.8\%}.
 \item[R2] \texttt{IF writing is excellent AND intelligence is high AND eloquence is high THEN profession is lawyer, \\
 confidence = 90\%, support = 1.8\%}.
\end{itemize}
}
Only based on the values of confidence and support, the two rules may seem equally plausible.  However, the rule predicting engineers is better than the rule predicting lawyers in that it correctly covers a higher percentage of instances belonging to the predicted class value.  In fact, Rule 1 covers and correctly classifies all 18 engineers in the dataset, while Rule 2 covers only 18 of the 72 lawyers. As follows from the base rate fallacy, even if the users know the base rates (or are presented with this information, e.g.,  as part of the exploratory data analysis), they will tend to not take full advantage of it when evaluating the plausibility of the presented rules. 
 
\paragraph{Debiasing techniques}
As a possible debiasing technique for the problem introduced above, users can also be presented the value of lift, which can be interpreted as an improvement over the base rate provided by the rule.  More precisely, lift is computed as a ratio of the confidence of the rule and a probability of the head of the rule, which corresponds to the base rate  (cf. also~Section~\ref{ss:decrule}).
The lift of Rule
1 is $0.9/0.018=50$, while lift of Rule 2 is $0.9/0.072=12.5$.
As can be seen, the value of lift is much higher for the first rule than for the second rule.
The reason is that the base rate for the engineer is 1.8\%, while for the lawyer profession it is 7.2\%, while the confidences of both rules are the same. For additional discussion of the use of lift as a debiasing technique, please refer to the debiasing part of Section~\ref{ss:wee} describing the weak evidence effect.

\citet{gigerenzer1995improve} show that representations in terms of natural frequencies, rather than conditional probabilities, facilitate the computation of cause's probability. 
Confidence is typically presented as a percentage in current software systems. The support rule quality metric is sometimes presented as a percentage and sometimes as a natural number. It would foster correct understanding if analysts are consistently presented natural frequencies in addition to percentages.  By adapting the diagrammatic representation of Bayes theorem ('Euler circles')  \cite{barbey2007base}, visualization of rules using diagrams presenting their coverage may also help reduce the base rate neglect. A specific proposal for debiasing techniques based on natural frequencies is present in Section~\ref{ss:reprh}, cf. also the discussion present in the following subsection.

\subsection{Insensitivity to Sample Size}
\label{ss:ins}
People tend to underestimate the increased benefit of higher robustness of estimates that are made on a larger sample, which is called insensitivity to sample size.
The insensitivity to sample size effect can be illustrated by the so-called hospital problem. In this problem, subjects are asked which hospital is more likely to record more days in which more than 60 percent of the newborns are boys. The options are a larger hospital, a smaller hospital, or both hospitals with about a similar probability. The correct expected answer---the smaller hospital---was chosen only by 22\% of participants in an experiment reported by \citet{Tversky27091974}. Insensitivity to sample size may be another bias resulting from use of the representativeness heuristic \citep{kahneman1972subjective}. When people use the representativeness heuristic, they compare the proportion of newborns who are boys to the proportion expected in the population, ignoring other relevant information. Since the proportion is similarly representative of the whole population for both hospitals, most of the participants believed that both hospitals are equally likely to record days in which more than 60 percent of the newborns are boys \citep{Tversky27091974}. However, these responses do not take into account that small samples have higher variance and therefore there is a higher  probability for recording an outlier not conforming to the underlying probability distribution. 

\paragraph{Differentiation from  the Base Rate Neglect} When determining the validity of assertions based on statistical evidence, insensitivity to sample size refers to the inability to appreciate the effect of sample size on the reliability of the provided estimate. In contrast, base rate neglect relates to the  tendency to not use the information about base rates, even though it is available.

\paragraph{Implications for rule learning}
This effect implies that analysts may be unable to appreciate the increased reliability of the confidence estimate with increasing the value of support, i.e., they may fail to appreciate that the strength of the connection between antecedent and consequent of a rule rises with an increasing the number of observations.
If confronted with two rules, where one of them has a slightly higher confidence and the second rule a higher support, this cognitive bias suggests that the analyst will prefer the rule with higher confidence (all other factors equal).

In the context of this bias, it is important to realize that population size is statistically irrelevant for the determination of sample size for large populations 
\citep{cochran2007sampling}.
However, previous research  \citep{bar1979role} has shown that the perceived sample accuracy can incorrectly depend on the sample-to-population ratio rather than on the absolute sample size. 
For a small population, a 10\% sample can be considered as more reliable than 1\% sample drawn from a much larger population.  This observation has substantial consequences for the presentation of rule learning results as support of a rule is typically presented as a percentage of the dataset size. Assuming that support relates to sample size 
and the number of instances in the dataset to population size, it follows that the presentation of support as a percentage (relative support)  induces the insensitivity to sample size effect.

To illustrate this effect on  interpretation of rule learning results by actual users, we will use data collected in experiment 3 reported in \cite{furnkranz2019cognitive}. In this experiment, participants were asked to evaluate pairs of rules in terms of plausibility. They were also presented  values of support and confidence in a frequency format for each of the rules. One pair of the evaluated rules had the following properties:
\begin{itemize}
    \item Rule 1: confidence 52.7\%, support 1.95\%, antecedent length 1
    \item Rule 2: confidence 52.1\%, support 25.9\%, antecedent length 1
\end{itemize}
Example justifications given by participants finding rule 1 plausible: ``confidence of test results slightly higher'', ``Both of these rules seem pretty terrible to me (53 and 52 percent are not impressive), but Rule 1 has a very marginally higher success rate.'', ``confidence is better''. Example justification for no preference was  ``both rules have similar probability of being more plausible''.
Interestingly, no participant evaluated rule 2 as more plausible in spite of  rule 2 having substantially higher support and very slightly higher confidence. The justifications provided by the participants refer only to the values of the interest measures, which shows that the actual semantics of the condition did not seem to have influenced their judgments. 
The overall results from this experiment indicate that if both confidence and support are explicitly revealed, confidence but not support will positively increase rule plausibility. A similar user study on a different dataset was performed in \cite{pagliaro2020cognitive} confirming our initial finding.

It follows that by increasing preference for higher confidence, this effect will generally contribute to a positive correlation between rule length and plausibility, since longer rules can better adapt to a particular group in data and thus have a higher confidence than more general, shorter rules. 
This is partly addressed by some state-of-the-art rule learning algorithms, which aim at learning short rules (cf. e.g., \cite{wang2017bayesian}). Such rules tend to be more general, have a higher support, and are thus statistically more reliable.

\paragraph{Debiasing techniques}
There have been successful experiments with providing decision aids to overcome the insensitivity to sample size bias. In particular, \citet{kachelmeier1990investigation} experimented with providing auditors a formula for computing appropriate sample size for substantive tests of details based on the description of a case and tolerable error. Provision of the aid resulted in larger sample sizes being selected by the auditors in comparison to intuitive judgment without the aid.
Similarly,  as the auditor  can choose the sample size, a user of an association rule learning algorithm can specify the minimum support threshold.
To leverage the debiasing strategy validated by \citet{kachelmeier1990investigation}, the rule learning interface should also inform the user of the effects of chosen support threshold on the accuracy of the confidence estimate of the resulting rules.  For algorithms and workflows where the user cannot influence the support of a discovered rule, relevant information should be available as a part of rule learning results. 
In particular, the value of rule support can be used to compute a confidence
interval for the value of confidence. Such supplementary information is already provided by Bayesian decision lists \citep{letham2015interpretable}, a recently proposed algorithmic framework positively evaluated with respect to interpretability (cf., e.g., \citep{de2017algorithmic}).
Another hypothesized debiasing method is to adopt the present support as an absolute number (absolute support), as demonstrated in Figure~\ref{fig:supp-conf} on page~\pageref{fig:supp-conf}. It should be noted that in the experiment reported in \cite{furnkranz2019cognitive}, a variation of the frequency format-based debiasing was used, but the insensitivity of sample size effect was still present. We therefore hypothesize that the frequency format supplementing the standard ratio presentation of support and confidence may reduce the incidence of this bias, but not completely remove it (cf. also discussion in Section~\ref{ss:reprh}). 

\subsection{Confirmation Bias and Positive Test Strategy}

Confirmation bias, sometimes also called myside bias, refers to the notion that people tend to look for evidence supporting the current hypothesis, disregarding conflicting evidence. According to \citet[p.\ 552]{evans1989bias} confirmation bias  is  ``the best  known  and  most widely  accepted  notion  of inferential  error  of human  reasoning.''\footnote{Cited according to \citet{nickerson1998confirmation}.} 

Research suggests that even neutral or unfavourable evidence can be interpreted to support existing beliefs, or, as \citet[p.\ 115--116]{trope1997wishful} put it, ``the same evidence can be constructed and reconstructed in different and even opposite ways, depending on the perceiver's hypothesis.'' 

A closely related phenomenon is the \emph{positive test strategy} (PTS) described by \citet{klayman1987confirmation}. This reasoning strategy suggests that when trying to test a specific hypothesis, people examine cases which they expect to confirm the hypothesis rather than the cases which have the best chance of falsifying it.  The difference between PTS and confirmation bias is that PTS is applied to test a candidate hypothesis while confirmation bias is concerned with hypotheses that are already established \citep[p.\ 93]{oswald2004confirmation}. 
The experimental results of \citet{klayman1987confirmation} show that under  realistic  conditions, PTS can be a very good  heuristic  for  determining  whether a hypothesis  is true or false, but it  can also lead  to systematic  errors if applied to an inappropriate task. 

\paragraph{Implications for rule learning}
This bias can have a significant impact depending on the purpose for which the rule learning results are used. If the analyst has some prior hypothesis before obtaining the rule learning results, according to the confirmation bias the analyst will tend to ``cherry pick'' rules confirming this prior hypothesis and  disregard  rules that contradict it. Given that some rule learners may output contradicting rules, the analyst may tend to select only the rules conforming to the hypothesis, disregarding applicable rules with the opposite conclusion,  which could otherwise turn out to be more relevant.

\paragraph{Debiasing techniques}
\label{ss:confbias-debias}

Delaying final judgment and slowing down work has been found to decrease confirmation bias in several studies \citep{spengler1995scientist,parmley2006effects}. User interfaces for rule learning should thus give the user not only the opportunity to save or mark interesting rules, but also allow the user to review and edit the model at a later point in time. An example rule learning system with this specific functionality is EasyMiner \citep{vojivr2018easyminer}.

\citet{wolfe2008locus} successfully experimented with  providing subjects with explicit guidelines for considering evidence both for and against a hypothesis. Provision of ``balanced'' instructions to search evidence for and against a given hypothesis reduced the incidence of confirmation bias  
from 50\% exhibited  by the control group to a significantly lower 27.5\%.
The assumption that educating users about cognitive illusions can be an effective debiasing technique for positive test strategy has been empirically validated on a cohort of adolescents by \citet{barberia2013implementation}.

Similarly, providing explicit guidance combined with modifications of the user interface of the system  presenting the rule learning results could also be considered. For example, in the context of explaining decisions of machine learning systems in the medical domain, \cite{wang2019designing} proposes to show prior probabilities of diagnoses. In the context of rule learning, a prior probability for the rule \texttt{IF} $A$ \texttt{AND} $B$ \texttt{THEN} $C$ is the (unconditional) probability $\Pr(C)$ of $C$ occurring in the data. This information is already made available by some rule learning systems, but not presented in a form that would be immediately understandable to the domain expert.

\begin{figure}[t]
\includegraphics[width=12cm]{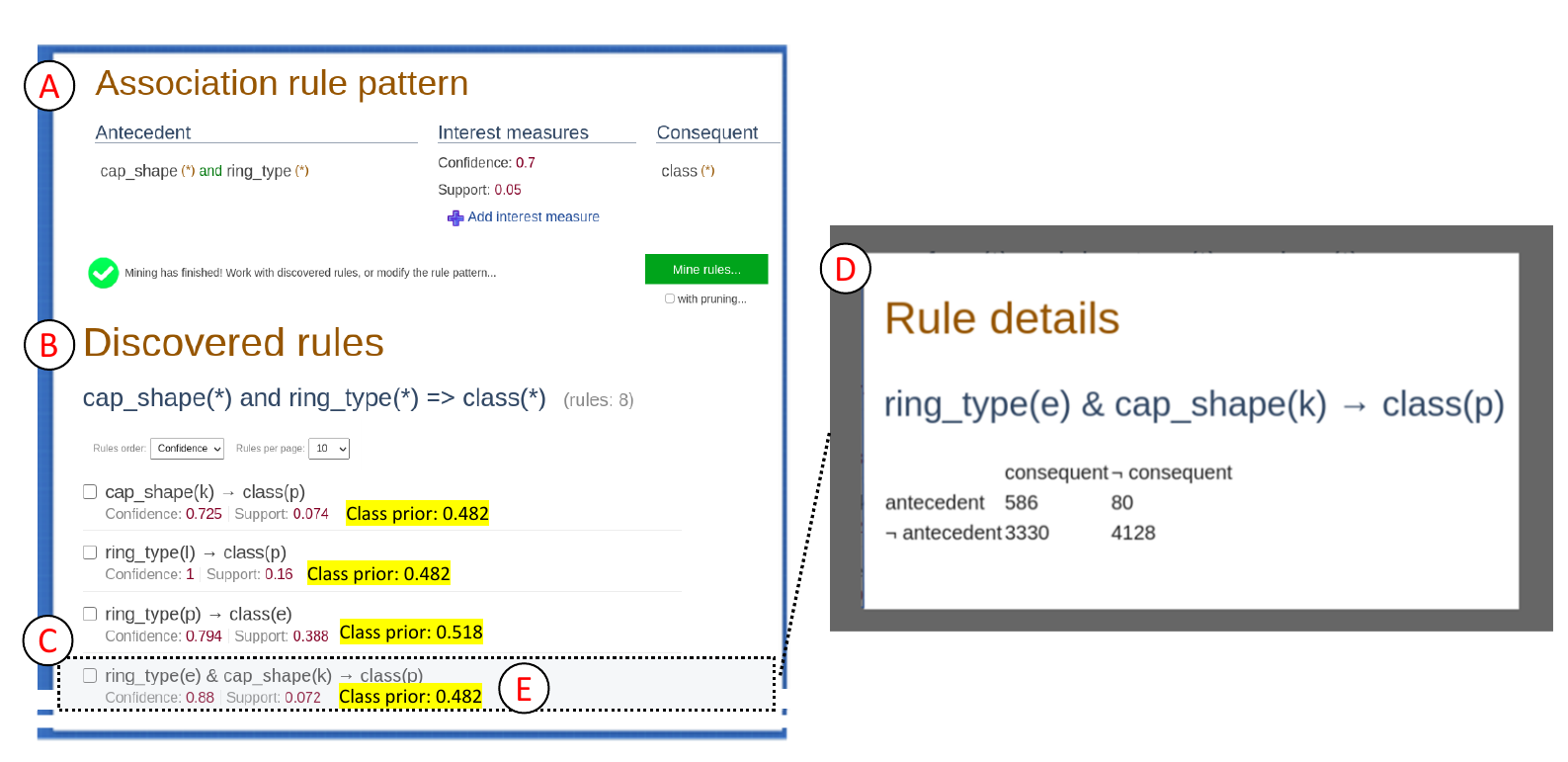}
\centering

\caption{Augmented screenshot from the EasyMiner rule learning system. A. User first sets a template to which the discovered rules must comply. Rules must also match the set minimum support and confidence thresholds; B. List of discovered rules; C. User chooses one rule; D. contingency table for chosen rule; E. (highlighted text) Proposed addition to the user interface -- inclusion of class priors. Note that while in the rest of the article a literal is denoted as, e.g., \texttt{cap\_shape=k}, the EasyMiner system represents it as  \texttt{cap\_shape(k)} following the notation of the GUHA method \cite{rauch2012observational}.} 

\label{fig:rulematrix}
\end{figure}

In Figure~\ref{fig:rulematrix} we illustrate this on an interface of the EasyMiner  web-based rule learning system \cite{vojivr2018easyminer}, which is featured as a representative of an open source tool used for education purposes (cf.~\cite{grossmann2015fundamentals}).

Most machine learning systems that allow rule learning output discovered rules in a form similar to what is shown in  Figure~\ref{fig:rulematrix} (bottom left), i.e., as  implications with associated values of confidence and support (both shown as percentages).

Although the information shown does not include the prior probabilities, some systems including EasyMiner allow showing a detailed contingency table for the selected rule (Figure~\ref{fig:rulematrix} right). This information is sufficient for the  user to compute the prior probabilities. For the selected rule with the class 'poisonous', this is $(586+3330)/8124=48.2$\%. 

As can be seen, even if supported by some systems, this process involves multiple steps and is inconvenient. We therefore suggest to include the prior probabilities into the primary description of the rule, alongside the main interest measures such as support and confidence. However, the contingency table has the advantage that the information it presents corresponds to the ``frequency format'', which is usually easier to understand. 

\subsection{Availability Heuristic}
The availability heuristic is a judgmental heuristic in which a person evaluates the frequency of classes or the probability of events by the ease with which relevant instances come to mind. This heuristic is explained by its discoverers, \citet{tversky1973availability}, as follows: ``That associative bonds are strengthened by repetition is perhaps the oldest law of memory known to man. The availability heuristic exploits the inverse form of this law, that is, it uses the strength of the association as a basis for the judgment of frequency.'' The availability heuristic is not itself a bias, but it may lead to biased judgments when availability is not a valid cue.

In one of the original experiments, participants were asked whether the letter ``R'' appears more frequently on the first or third position in English texts \citep{tversky1973availability}. About 70\% of participants answered incorrectly that it appears more frequently on the first position, presumably because they estimated the frequency by recalling words containing ``R'' and it is easier to recall words starting with R than words with R on the third position.

While original research did not distinguish between the number of recollected instances and ease of the recollection, later studies showed that to determine availability, it is sufficient to assess the ease with which instances or associations could be brought to mind; it is not necessary to count all the instances one is able to come up with \citep{schwarz1991ease}. 

\paragraph{Implications for rule learning}
An application of the availability heuristic in rule learning would be based on the ease of recollection of instances (examples) matching the \emph{complete}  rule (all conditions and consequent) by the analyst.
Rules containing conditions for which instances can be easily recalled would be found more plausible compared to rules not containing such conditions. As an example, consider the rule pair

\medskip
\begin{tabular}{cl}
$R_1$: & \texttt{IF latitude $\leq$ 44.189 AND longitude $\leq$  6.3333} \\ &\texttt{AND longitude $>$  1.8397 THEN Unemployment is high}\\ 
$R_2$: & \texttt{IF population $\leq$ 5 million THEN Unemployment is high}.
\end{tabular}
\medskip

\noindent
It is arguably easier to recall specific countries matching the second rule, than countries matching the conditions of the first rule.

It is conceivable that the availability heuristic could also be applied in case when the easily recalled instances match  \emph{only some of the conditions} in the antecedent of the rule, such as only latitude in the example above.  The remaining conditions would be ignored.

On the other hand, such a bias can also be implemented as a bias into rule learning algorithms. Often, in particular in cases where many candidate conditions are available, such as datasets with features derived from the semantic web \citep{Ristoski2016}, the same information can be encoded in rules that use different sets of conditions. For example, \citet{gabriel2014learning} proposed an algorithm that gives preference to selecting conditions that are semantically coherent. A similar technique could be used for realizing a preference for attributes that are easier to recall for human analysts.

\paragraph{Debiasing techniques}
Several studies have found that people use the ease of recollection in judgment only when they cannot attribute it to an irrelevant cause \citep{schwarz2004metacognitive}. Alerting an analyst to the reason why instances matching the conditions in the rule under consideration are easily recalled should, therefore, reduce the impact of the availability heuristic as long as the reason is deemed irrelevant to the task at hand.

\subsection{Reiteration Effect, Effects of Validity and Illusiory Truth}
\label{ss:reiteration}
The reiteration effect describes the phenomenon that repeated statements tend to become more believable  \citep{hertwig1997reiteration,pachur2011recognition}.
For example, in one experiment, \citet{hasher1977frequency} presented subjects with general statements and asked them to assess their validity. Some of the statements were false and some were true.  The experiment was conducted in several sessions, where some of the statements were repeated in subsequent sessions.  The average perceived validity  of both true and false repeated statements  rose between the sessions, while for non-repeated statements it dropped slightly.

The effect is usually explained by use of the processing fluency in judgment. Statements that are processed fluently (easily) tend to be judged as true and repetition makes processing easier. A recent alternative account argues that repetition makes the referents of statements more coherent and people judge truth based on coherency \citep{unkelbach2017referential}.

The reiteration effect is also known under different labels, such as ``frequency-validity'' or ``illusory truth'' \citep[195]{hertwig1997reiteration}. However, some research suggests that these are not identical phenomena.
For example, the \emph{truth effect} ``disappears when the actual truth status is known''
\citep[p.~253]{pohl2017cognitive}, which does not hold for validity effect in general.

\paragraph{Implications for rule learning}
In  the rule learning context, a repeating statement which becomes more believable corresponds to the entire rule or possibly a ``subrule'' consisting of the consequent of the rule and a subset of conditions in its antecedent. A typical rule learning result contains multiple rules that are substantially overlapping. If the analyst is exposed to multiple similar statements, the reiteration effect will increase the analyst's belief in the \emph{repeating} subrule. 
Especially in the area of association rule learning, a very large set of redundant rules---covering the same, or nearly same set of examples---is routinely included in the output. 

\citet{schwarz2007metacognitive} suggest that a mere 30 minutes of delay can be enough for information originally seen as negative to have a positive influence.
Applying this in a data exploration task, consider an analyst who is presented a large number of ``weak'' rules  corresponding to highly speculative patterns of data. Even if the analyst  rejects the rule---for example based on the presented metrics, pre-existing domain knowledge or common sense---the reiteration effect will make the analyst  more prone to accept a similar rule later.

\paragraph{Debiasing techniques}
The reiteration effect can be suppressed already on the algorithmic level by ensuring that rule learning output does not contain redundant rules. This can be achieved by pruning algorithms  \citep{furnkranz1997pruning}. Another possible technique is presenting the result of rule learning in several layers, where only clusters of  rules (``rule covers'') summarizing multiple subrules are presented at first \citep{ordonez2006constraining}. The user can expand the cluster  to obtain more similar rules. A more recent algorithm that can be used for summarizing multiple rules is the meta-learning method proposed by \citet{berka2018comprehensive}. 

Several lessons can be learnt from \citet{hess2006psychological}, who studied the role of the reiteration effect for spreading of gossip. Interestingly, already simple reiteration was found to increase perceived gossip veracity, but only for those who found the gossip relatively uninteresting.  Multiple sources of gossip were found to increase its perceived veracity, especially when these sources were independent. Information that explained the gossip by providing benign interpretation  decreased the perceived veracity of gossip. 
These findings suggest that it is important to explain to the analyst which rules share the same source, i.e. what is the overlap in their coverage in terms of specific instances. Second, explanations can be improved by utilisation of recently proposed techniques that use domain knowledge to  filter or explain rules, such as expert deduction rules proposed by \citet{Rauch2018}.

The research related to debiasing the reiteration effect has been largely centered around the problem of debunking various forms of misinformation (cf., e.g., \citep{schwarz2007metacognitive,lewandowsky2012misinformation,ecker2017reminders}). The current largely accepted recommendation is that to correct misinformation, it is best to address it directly -- 
repeat the misinformation along with arguments against it \citep{lewandowsky2012misinformation,ecker2017reminders}.
This can be applied, for example, in incremental machine learning settings, when the results of learning are revised when new data arrive, or when mining with formalized domain knowledge. Generally, when the system has knowledge of the analyst being previously presented a rule (a hypothesis), which is falsified following the current state of knowledge, the system can explicitly notify the analyst, listing the rule in question and explaining why it does not hold.

\subsection{Mere Exposure Effect}
\label{ss:mereexp}
According to the mere exposure effect, repeated exposure to an object results in an increased preference (liking, affect)  for that object. When  a  concrete stimulus  is  repeatedly  exposed, the preference  for  that  stimulus  increases logarithmically as  a
function  of  the  number  of  exposures \citep{bornstein1989exposure}. The size of the mere exposure effect also depends on whether the stimulus the subject is exposed to is exactly the same as in prior exposure or only similar to it \citep{monahan2000subliminal}---the same stimuli are associated with a  larger mere exposure effect. 

The mere exposure effect is another consequence of increased fluency of processing associated with repeated exposure (cf.\ Section~\ref{ss:reiteration}) \citep{winkielman2003hedonic}.

Duration of the exposure below 1 second produces the strongest mere exposure effect, with increasing time of exposure the effect drops and  repeating exposures decrease the effect. The liking induced by the effect drops more quickly with increasing exposures when the presented stimulus is simple (e.g., an ideogram) as opposed to complex (e.g., a photograph)  \citep{bornstein1989exposure}. A recent meta-analysis suggests that there  is an inverted-U
shaped  relation  between  exposure
and affect \citep{montoya2017re}.

\paragraph{Differentiation from the reiteration effect}
While the reiteration effect referred to the use of processing fluency in the judgment of truth, the mere exposure effect relates to the positive feeling that is associated with fluent processing. The mere exposure effect, unlike the reiteration effect \citep[p.~245]{dechene2010truth}, has been also found to depend on the duration of stimulus exposure.

\paragraph{Implications for rule learning}
The  extent to which the mere exposure effect can affect the interpretation of rule learning results is limited by the fact that its magnitude decreases with extended exposure to the stimuli. It can be expected that the analysts inspect the rule learning results for a much longer period of time than the 1 second below which exposure results in the strongest effects \citep{bornstein1989exposure}.  However, it is not unusual for rule-based models to be composed of several thousand rules \citep{alcala2011fuzzy}. When the user scrolls through a list of rules, each rule can be shown only for a fraction of a second. The analyst is not aware of having seen the rule, yet the rule can influence the analyst's judgment through the mere exposure effect.

The mere exposure effect can also play a role when rules from the text mining or sentiment analysis domains are interpreted. 
The initial research of the mere exposure effect by \citet{zajonc1968attitudinal} included experimental evidence on the positive correlation between word frequency and affective connotation of the word. From this, it follows that a rule containing frequently occurring words can induce the mere exposure effect. 

\paragraph{Debiasing techniques}
While there is a considerable body of research focusing on the mere exposure effect, our literature survey  did not result in any directly applicable debiasing techniques. Only recently, \citet{becker2016reversing} reported the reversal of the mere exposure effect when people fearful of spiders were presented spider pictures. Contrary to the mere exposure effect, the participants preferred pictures of spiders they had not seen before to those already seen. 
This result, although interesting, is difficult to transpose to the domain of rules given that it depends on the specific characteristic of the people and evaluated objects. 

Nevertheless, there are some conditions known to decrease the mere exposure effect that can be utilized in  machine learning  interfaces. The effect is strongest for repeated,  ``flash-like'' presentation of information. A possible workaround is to avoid subliminal exposure completely, by changing the mode of operation of the corresponding user interfaces. One attempt at a user interface to rule learning respecting these principles is the EasyMiner system \citep{easyminer12,vojivr2018easyminer}, in which the user can precisely formulate the mining task as a query against data. This restricts the number of rules that are discovered and the user is consequently exposed to.

\subsection{Overconfidence and underconfidence}
\label{ss:effdif}
A decision maker's judgment is normally associated with the belief that the judgment is true, i.e., with confidence in the judgment. \citet{griffin1992weighing} argue that confidence in judgment is based on a combination of the strength of evidence and its weight (credibility).
According to their studies, people tend to combine strength with weight in suboptimal ways, resulting in the decision maker being too much or too little confident about the hypothesis at hand than would be normatively appropriate given the available information. This discrepancy between the normative confidence and the decision maker's confidence is called \emph{overconfidence} or \emph{underconfidence}.

People use the provided data to assess a hypothesis, but they insufficiently regard  the quality of the data. 
\citet{griffin1992weighing} describe this manifestation of bounded rationality as follows: ``If people focus primarily on the warmth of the recommendation  with insufficient regard for  the credibility of the writer, or the correlation between the predictor and the criterion, they will be overconfident when they encounter a glowing letter based on a casual contact, and  they will be underconfident when they encounter a moderately positive letter from a highly knowledgeable source.''

\paragraph{Implications for rule learning}
Research has revealed systematic patterns of overconfidence and underconfidence \citep[p.~426]{griffin1992weighing}: If the estimated difference between two competing hypotheses is large, it is easy to say which one is better and there is a pattern of underconfidence. As the degree of difficulty rises (the difference between the normative confidence of two hypotheses is decreasing), there is an increasing pattern of overconfidence.

The strongest overconfidence was recorded for problems where the \emph{weight of evidence is low and the strength of evidence is high}. This directly applies to rules with a high value of confidence and low value of support. 
The empirical results related to the effect of difficulty, therefore, suggest that the predictive ability of such rules will be substantially overrated by analysts. This is particularly interesting because rule learning algorithms often suffer from a tendency to unduly prefer overly specific rules that have a high confidence on small parts of the data to more general rules that have a somewhat lower confidence, a phenomenon also known as overfitting. The above-mentioned results seem to indicate that humans suffer from a similar problem (albeit presumably for different reasons), which, e.g., implies that a human-in-the-loop solution may not alleviate this problem.

\paragraph{Debiasing techniques}
Research applicable to debiasing of overconfidence originated in 1950', but most initial efforts to reduce overconfidence  have failed \citep{fischoff1981debiasing,arkes1987two}. 
Some recent research focuses on the hypothesis that the feeling of confidence reflects  factors indirectly related to choice processes \citep{fleisig2011adding,hall2007illusion}. For example,  in a sports betting experiment performed by   \citet{hall2007illusion}, participants underweighted  statistical cues while betting when they knew the names of players. This research leads to the conclusion that ``more knowledge can decrease accuracy and simultaneously increase prediction confidence'' \citep{hall2007illusion}. Applying this to debiasing in the rule learning context, presenting less information can be achieved by reducing the number of rules and removing some conditions in the remaining rules. This can be achieved by a number of methods, such as feature selection or an external setting of maximum antecedent length, which is permitted by some (but not all) implementations of rule learning algorithms.  Also, rules and conditions that do not pass a statistical significance test can be removed from the output. 

As with other biases,  research on debiasing  overconfidence points at the importance of educating the experts on principles of subjective probability judgment and the associated biases \citep{clemen2002debiasing}.
\citet[p. 487]{shafir2013behavioral} recommends to debias overconfidence (in policymaking) by making the subject hear both sides of an argument. 
 In the rule learning context, this would correspond to the user interface making rules and knowledge easily accessible, which  is in ``unexpectedness'' or ``exception'' relation with  the rule in question, as, e.g., experimented with in  frameworks  postprocessing association rule learning results \citep{Kliegr:2011:SSA:2070639.2070657}.

\subsection{Recognition Heuristic}
\citet{pachur2011recognition} define the recognition heuristic as follows:  ``For two-alternative choice tasks, where one has to decide which of two objects scores higher on a criterion, the heuristic can be stated as follows: If one object is recognized, but not the other, then infer that the recognized object has a higher value on the criterion.'' In contrast with the availability heuristic, which is based on ease of recall, the recognition heuristic is based only on the fact that a given object is recognized. However, the two heuristics could be combined. When only one object in a pair is recognized, then the recognition heuristic would be used for judgment. If both objects are recognized, then the speed of the recognition could influence the choice \citep{hertwig2008fluency}.

The use of this heuristic could be seen from an experiment performed by \citet{goldstein1999recognition}, which focused on estimating which of two cities in a presented pair is more populated. People using the recognition heuristic would say that the city they recognize has a higher population. The median proportion of judgments complying to the recognition heuristic was 93\%. It should be noted that the application of this heuristic  is in this case ecologically justified since recognition will be related to how many times the city appeared in a newspaper report, which in turn is related to the city size \citep{beaman2006does}.

\paragraph{Implications for rule learning}
The recognition heuristic can manifest itself by a preference for rules containing a recognized  attribute name or value in the antecedent of the rule. 
Analysts processing rule learning results are typically shown many rules, contributing to time pressure. This can further increase the impact of the recognition heuristic.

Empirical results reported by \citet{michalkiewicz2018smarter} indicate that people with higher cognitive ability 
use the recognition heuristic  more when it is successful and less when it is not. The work of \citet{pohl2017use} shows that people  adapt their decision strategy with respect to the more general environment rather than the specific items they are faced with. Considering that the application of the recognition heuristic can in some situations lead to better results than the use of available knowledge, the recognition heuristic may not necessarily have overly negative impacts on the interpretation of rule learning results.

In feedback collected as part of experiments conducted in \cite{furnkranz2019cognitive} there is some evidence of the recognition heuristic having been applied. 
For the mushroom dataset, there is a number of responses where the anise smell is explicitly given as a reason for positive preference on the grounds of prior experience (recognition): ``Anise smells nice and usually nice smelling things aren't poisonous'', ``personal experience''. Similar observations have been made for another dataset, where the participants were asked to say which of two presented rules predicting quality of living in a city is more plausible. The following justification for choosing a particular rule --  ``Cities in Switzerland tend to be nice'' -- bears close resemblance to the general psychological experiment designed by \cite{goldstein1999recognition} cited above, in which participants were asked which city was larger.

\paragraph{Debiasing techniques}
Under time pressure people assign a higher value to recognized objects than to unrecognized objects. This happens also in situations when recognition is a poor cue \citep{pachur2006psychology}. 
Changes to user interfaces that induce ``slowing down'' could thus help to address this bias.
As to the alleviation of effects of recognition heuristic in situations where it is ecologically unsuitable, \citet{pachur2006psychology} note that suspension of the heuristic requires additional time or direct knowledge of the ``criterion variable''. 
In typical real-world machine learning tasks, the data can include a high number of attributes that even experts are not acquainted with in detail. When these are recognized (but not understood), 
even experts may be liable to the recognition heuristic. When information on the meaning of individual attributes and literals is made easily accessible, we conjecture that the application of the recognition heuristic can be suppressed.

\subsection{Information Bias }
\label{ss:information}
Information bias refers to the tendency to seek more information to improve the perceived validity of a statement even if the additional information is not relevant or helpful. 
The typical manifestation of the information bias is evaluating questions as worth asking even when the answer cannot affect the hypothesis that will be accepted \citep{baron1988heuristics}. 

For example, \citet{baron1988heuristics} asked subjects to  assess to what degree a medical test is suitable for deciding which of three diseases to treat. The test detected a chemical, which was with a certain probability associated with each of the three diseases. These probabilities varied across the cases. Even though in some of the cases an outcome of the test would not change the most likely disease and thus the treatment, people tended to judge the test as worth doing. 
While information bias is primarily researched in the context of information acquisition \citep{nelson2010experience,nelson2005finding}, some scientists interpret this more generally as judging features with zero probability gain as useful, having potential to change one's belief \citep[p. 158]{nelson2008towards}.

\paragraph{Implications for rule learning}
Many rule learning algorithms allow the user to select the size of the generated model -- in terms of the number of rules that will be presented, as well as by setting  the maximum length of conditions of the generated rules. Either as part of the feature selection or when defining constraints for the learning the users decide which attributes are relevant. These can then appear among conditions of the discovered rules. 

According to the information bias, people will be prone to set up the task so that they receive more information -- resulting in a larger rule list with longer rules containing attributes with little information value. 

It is unclear if the information effect applies also to the case when the user is readily presented with more information, rather than given the possibility to request more information. Given the proximity of these two scenarios, we conjecture that information bias (or some related bias) will make people prefer more information to less, even if it is obviously not relevant. 
This conjecture is supported by the  feedback collected as part of experiments conducted in \cite{furnkranz2019cognitive}. When choosing which  of the two presented rules is more plausible, multiple participants have chosen the longer rule providing justification such as:  ``more describing factors make identification easier'', ``more indicators means more certainty'', ``Rule 1 has a much tighter definition of what would constitute a poisonous mushroom with 5 conditions as compared to rule 2 which only contains just 1 condition for the same result, so rule 1 has a much higher plausibility of being believable''.   

\paragraph{Debiasing techniques}
While informing people about the diagnosticity of the considered questions does not completely remove the information bias, it reduces it \citep{baron1988heuristics}.
To this end, communicating attribute importance can help guide the analyst in the task definition phase.

Although existing algorithms and systems already provide ways for determining the importance of individual rules, for example via values of confidence, support, and lift, the cues on the importance of individual conditions in rule antecedent are typically not provided. While feature importance is  computed within many learning algorithms, it is often used only internally. Exposing this information to the user can help counter the information bias. 

\subsection{Ambiguity Aversion }
\label{ss:ambiguity}
Ambiguity aversion refers to the tendency to prefer known risks over unknown risks. This is often illustrated by the Ellsberg paradox \citep{ellsberg1961risk}, which shows that humans
tend to systematically prefer a bet with a known probability of winning over a bet with a not precisely known probability of winning, even if it means that their choice is systematically influenced by irrelevant factors.

As argued by \citet{Camerer1992}, ambiguity aversion is related to the information bias: the demand for information in cases when it has no effect on the decision can be explained by the aversion to ambiguity --- people dislike having missing information. That is, the information bias can be seen as a possible consequence of ambiguity aversion in the case when there is a possibility to seek additional information.

\paragraph{Implications for rule learning}
The ambiguity aversion may have profound implications for rule learning. The typical data mining task will contain a number of attributes the analyst has no or very limited knowledge of. The ambiguity aversion will manifest itself in a preference for rules that do not contain ambiguous conditions.

In feedback collected as part of experiments conducted in \cite{furnkranz2019cognitive},  arguments were given as reasons for selecting the alternative rule not containing unknown value as more plausible. Specifically, in the Mushroom dataset, one rule of the two presented rules was considered more plausible because the other rule contained an unknown value or trait: ``Could a mushroom smell of creosote?'', ``It's hard to tell what creosote smells like.''
In this context, the recognition heuristic  can be associated with making risk-averse decisions which is one of the  hypothesized etiologies of cognitive biases (cf. Section~\ref{sec:functions}). In the example above, conditions not recognized (such as creosote), were considered to be predictive of the worse outcome (poisonous mushroom). 

\paragraph{Debiasing techniques}
An empirically proven way to reduce ambiguity aversion is accountability -- ``the expectation on the side of the
decision maker of having to justify her decisions to somebody else'' \citep{vieider2009effect}. This debiasing technique is hypothesized to work through higher cognitive effort that is induced by accountability.

This can be applied in the rule learning context by requiring the analysts to provide  justifications for why they evaluated a specific discovered rule as interesting. Such an explanation can  be textual, but also can have a structured form.
To decrease demands on the analyst, the explanation may only be required if a conflict with existing knowledge has been automatically detected, for example, using the approach based on the deduction rules as proposed by \citet{Rauch2018}. 

Since the application of the ambiguity aversion can partly stem from the lack of knowledge of the conditions included in the rule, it is conceivable that this bias would be alleviated if the description of the meaning of  the conditions  is made easily accessible to the analyst, as demonstrated in  e.g. \citep{Kliegr:2011:SSA:2070639.2070657}.

\subsection{Confusion of the Inverse }
\label{ss:confusioninverse}
This effect corresponds to confusing the probability of cause and effect, or, formally,
the confidence of an implication $A \rightarrow B$ with its inverse $B \rightarrow A$, i.e., $\Pr(B \mid A)$ is confused with the inverse probability $\Pr(A \mid B)$. For example,  \citet{villejoubert2002inverse} showed in an experiment that about half of the participants estimating the probability of membership in a class gave most of their estimates that corresponded to the inverse probability.

\paragraph{Implications for rule learning}
The confusion of the direction of the IF-THEN implication sign has significant consequences on the interpretation of a rule. 
Already
\citet{michalski1983theory}  noted that there are two different kinds
of rules, discriminative and characteristic. 
\emph{Discriminative rules} can quickly discriminate an
object of one category from objects of other categories. A simple
example is the rule 
$$\texttt{IF trunk THEN elephant}$$
which states that an animal with a trunk is an elephant. This implication
provides a simple but effective rule for recognizing elephants
among all animals.

\emph{Characteristic rules}, on the other hand, try to capture \emph{all}
properties that are common to the objects of the target class. A rule
for characterizing elephants could be 
$$\texttt{IF elephant THEN heavy, large, grey, bigEars, tusks, trunk.}$$
Note that here the implication sign is reversed: we list all
properties that are implied by the target class, i.e., by an animal
being an elephant. From the point of understandability, characteristic
rules are often preferable to discriminative rules. For example, in a
customer profiling application, we might prefer to not only list a few
characteristics that discriminate one customer group from the other,
but we are rather interested in all characteristics of each customer group.

Characteristic rules are very much related to \emph{formal concept
  analysis} \citep{FCA,FCA-Foundations}. Informally, a concept is
defined by its intent (the description of the concept, i.e., the
conditions of its defining rule) and its extent (the instances that
are covered by these conditions). A \emph{formal concept} is then a
concept where the extension and the intension are Pareto-maximal, i.e.,
a concept where no conditions can be added without reducing the number
of covered examples. In Michalski's terminology, a formal concept is
both discriminative and characteristic, i.e., a rule where the head is
equivalent to the body.

This confusion may manifest itself strongest in the area of association rule learning, where an attribute can be of interest to the analyst both in the antecedent and consequent of a rule.

\paragraph{Debiasing techniques}
The confusion of the inverse seems to imply that humans will not clearly distinguish between these types of rules, and, in particular, tend to interpret an implication as an equivalence. From this, we can infer that characteristic rules, which add all possible conditions even if they do not have additional discriminative power, may be preferable to short discriminative rules.

In terms of generally applicable debiasing techniques, \citet{edgell2004learned} studied the influence of the effect of training of analysts in probabilistic theory with the conclusion that it is not effective in addressing the confusion of the inverse fallacy.

\citet[p.~195]{werner2018eliciting} point at a concern regarding the use of language liable to misinterpretation in  statistical textbooks teaching fundamental concepts such as independence. The authors illustrate the misinterpretation on the statement \emph{whenever Y has
no effect on X} as ``This statement is used to explain that two variables, X and Y, are independent and their joint distribution is simply the product of their margins. However, for many experts, the term 'effect' might imply a causal relationship.'' 
From this, it follows that representations of rules should strive for an unambiguous meaning of the wording of the implication construct. The specific recommendations provided by \citet{diaz2010teaching} for teaching probability  can also be considered in the next generation of textbooks aimed at the data science audience.

\subsection{Context and Tradeoff Contrast Effects}
People evaluate objects in relation to other available objects, which may lead to various effects of the context of the presentation of a choice.
For example, in one of the experiments described by \citet{tversky1993context}, subjects were asked to choose between two microwave ovens: Panasonic and Emerson. The number of subjects who chose Emerson was 57\% and 43\% chose Panasonic. Another group of subjects was presented the same problem with the following manipulation: A more expensive, but inferior Panasonic was added to the list of possible options. After this manipulation, only 13\% chose the more expensive Panasonic, but the number of subjects choosing the cheaper Panasonic rose from 43\% to 60\%. That is, even though the additional option was dominated by the cheaper Panasonic device and it should have been therefore irrelevant to the relative preference of the other ovens, its addition changed the preference in favour of the better Panasonic device.
The experiment thus shows that selection of one of the available alternatives, such as products or job candidates,  can be manipulated by addition or deletion of alternatives that are otherwise irrelevant.
\citet{tversky1993context} attribute the tradeoff effect to the fact that ``people often do not have a global preference order and, as a result, they use the context to identify the most 'attractive' option.''

It should be noted that according to \citet{tversky1993context} if people have well-articulated preferences, the background context has no effect on the decision.

\paragraph{Implications for rule learning}
The  effect could be illustrated on the inter-rule comparison level. In the base scenario, a constrained rule learning yields only a rule $R_1$ with a confidence value of $0.7$. Due to the relatively low value of confidence, the user does not find the rule very plausible. By lowering the minimum confidence threshold, multiple other rules predicting the same target class are discovered and shown to the user. These other rules, inferior to $R_1$, would increase the plausibility of $R_1$ by the tradeoff contrast effect.

\paragraph{Debiasing techniques}
Marketing professionals sometimes introduce more expensive versions of the main product, which induces the tradeoff contrast. The presence of a more expensive alternative with little added value increases sales of the main product \citep{simonson1992choice}. Somewhat similarly, a rule learning algorithm can have on its output rules with very high confidence, sometimes even 1.0, but very low values of support, or rules with irrelevantly low confidence, but high support. Removal of such rules can help to debias the analysts.

The influence of context can in some cases improve communication \citep[p. 293]{simonson1992choice}.
An attempt at making contextual attributes explicit in the rule learning context was made by \citet{SD-CHD}, who introduced \emph{supporting factors} as a means for complementing the explanation delivered by conventional learned rules. Essentially, supporting factors are additional attributes that are not part of the learned rule, but nevertheless have very different distributions with respect to the classes of the application domain. 
In line with the results of \citet{NB-Rules-Application}, medical experts found that these supporting factors increase the plausibility of the found rules.

\subsection{Negativity Bias}
According to the negativity bias, 
negative evidence 
tends to have a greater effect than neutral or positive evidence of equal intensity \citep{rozin2001negativity}. 

For example, the experiments by \citet{pratto2005automatic} investigated whether the valence of a word (desirable or undesirable trait)  has effect on the time required to identify the colour in which the word appears on the screen. The results showed that the subjects took longer to name the colour of an undesirable word than for a desirable word. The authors argued that the response time was higher for undesirable words because undesirable traits get more attention.
Information with negative valence is given more attention partly because people seek diagnostic information, and negative information is more diagnostic \citep{skowronski1989negativity}.
Some research suggests that negative information is better memorized and subsequently recognized \citep{robinson1996role,ohira1998effects}. 

\paragraph{Implications for rule learning}

An interesting applicable discovery shows that negativity is an ``attention magnet'' \citep{fiske1980attention,ohira1998effects}. This implies that a rule predicting a class phrased with negative valence will get more attention than a rule predicting a class phrased with words with positive valence. 

We expect this effect to apply also to conditions in the antecedent of the rules.  In experiments conducted in \cite{furnkranz2019cognitive}, we presented participants a rule that contained conditions with clearly negative valence ('foul' smell) and a rule containing multiple  neutral conditions, but also often associated with poisonous mushrooms.

\medskip
\begin{tabular}{p{0.2cm}p{11cm}}
 $R_1$: & \texttt{IF veil color is  white and   gill spacing is close and mushroom does not have bruises and mushroom has one ring THEN the mushroom is poisonous
}\\ 
$R_2$: & \texttt{IF odour is foul THEN the mushroom is poisonous}.
\end{tabular}
\medskip

All participants who provided the rationale for their plausibility judgment referred to the foul smell, even though some of them preferred $R_1$. This  could illustrate how the negativity bias could influence analysts, even though in this example the observation is not necessarily a proof of the bias.

\paragraph{Debiasing techniques}
 Putting a higher weight to negative information may in some situations be a valid heuristic. What needs to be addressed are cases, when the relevant piece of 
information (a condition in the rule or the predicted class) is positive and a less relevant piece of information is negative \citep{huber2010mindless,tversky1981framing}. It is therefore advisable that any such suspected cases are detected in the data preprocessing phase, and the corresponding attributes or values are replaced with more neutral-sounding alternatives.

\subsection{Primacy Effect}
Once people form an initial assessment of plausibility  (favourability) of an option, its subsequent evaluations will reflect this initial disposition. 

\citet{bond2007information} investigated to what extent changing the order of information which is presented to a potential buyer affects the propensity to buy. For example, in one of the experiments, if the positive information (product description) was presented as first, the number of participants indicating they would buy the product was 48\%. When the negative information (price) was presented first, this number decreased to 22\%. \citet{bond2007information} argue that the effect is caused by the distortion of interpretation of new information in the direction of the already held opinion. The information presented first not only influences disproportionately the final opinion, but it also influences the interpretation of novel information.

\paragraph{Implications for rule learning}
Following the primacy effect, the analyst will favour rules that are presented as first in the rule model. Largest negative effects of this bias are likely to occur when such ordering does not correspond to the quality of the rules (as e.g., reflected by confidence, support or lift), for example, when rules are presented in the order in which they were discovered by a breadth-first algorithm.
In this case, \emph{mental contamination} is another applicable bias related to the primacy effect (or in general order effects). This refers to the case when  a presented hypothesis can influence subsequent decision making by its content, even if the subject is fully aware of the fact that the presented information is purely speculative \citep{fitzsimons2001nonconscious}. Note that our application scenario differs from \citep{fitzsimons2001nonconscious} and some other related research, in that cognitive psychology mostly investigated the effect of \emph{asking a hypothetical question}, while we are concerned with considering the plausibility of a presented hypothesis (inductively learnt rule).
\citet{fitzsimons2001nonconscious} found that respondents are generally not able to prevent the contamination effects of the hypothetical questions and that the bias increases primarily when the hypothetical question is relevant. This bias is partly attributed to the application of expectations related to conversational maxims \citep{gigerenzer1999overcoming}.

\paragraph{Debiasing techniques}
Three types of debiasing techniques were examined by \citet{mumma1995procedural} in the context of clinical-like judgments. The \emph{bias inoculation}  intervention involves direct training on the applicable bias or biases, consisting of information on the bias, strategies for adjustment, as well as completing several practical assignments. The second technique was \emph{consider-the-opposite} debiasing strategy, which sorts the  information  according  to  diagnosticity before it is reviewed. The third strategy evaluated was simply \emph{taking notes} when reviewing each cue before the final judgment was made. Interestingly, bias inoculation, a representative of direct debiasing techniques, was found to be the least effective. Consider-the-opposite and taking notes were found to work equally well. 

To this end, a possible debiasing strategy can be founded in the presentation of the most relevant rules first. Similarly, the conditions within the rules can be ordered by predictive power.
Some rule learning algorithms,  such as CBA \citep{Liu98integratingclassification},  partly take advantage of the primacy effect, since they create rule models that contain rules sorted  by their strength. CBA sorts rules by confidence, then by support and rule length
Other algorithms order rules so that more general rules (i.e., rules that cover more examples) are presented first. This typically also corresponds to the order in which rules are learned with the commonly used separate-and-conquer or covering strategies \cite{furnkranz1999separate}. Simply reordering the rules output by these algorithms may not work in situations, when rules compose a rule list that is automatically processed for prediction purposes.\footnote{One technique that can positively influence comprehensibility of the rule list is  prepending (adding to the beginning) a new rule to the previously learned rules \citep{Prepend}. The intuition behind this argument is that there are often simple rules that would cover many of the positive examples, but also cover a few negative examples that have to be excluded as exceptions to the rule. Placing the simple general rule near the end of the rule list allows us to handle exceptions with rules that are placed before the general rule and keep the general rule simple.}
In order to take advantage of the note-taking debiasing strategy, the user interface can support the analyst in annotating the individual rules. 

\citet{lau2009can} provide a reason for optimism concerning the debiasing effect stemming from the proposed changes to user interface of machine learning tools. Their paper showed a debiasing effect of similar changes implemented in a user interface of an information retrieval system used by consumers to find health information. Three versions of the system were compared: a baseline ``standard'' search interface, \emph{anchor debiasing interface}, which asked the users to annotate the read documents  as providing evidence for/against/neutral the proposition in question. Finally, the \emph{order debiasing interface} reordered the documents to neutralize the primacy bias by creating a ``counteracting order bias''. This was done by randomly reshuffling a part of the documents. When participants used the baseline and anchor debiasing interface, the order effect was present. 
On the other hand, the use of the order debiasing interface eliminated the order effect \citep{lau2009can}.

\subsection{Weak Evidence Effect}
\label{ss:wee}
According to the weak evidence effect, presenting weak evidence in favour of an outcome can actually decrease the probability that a person assigns to the outcome. For example, in an experiment in the area of forensic science reported by \citet{martire2013expression}, it was shown that participants presented with evidence weakly supporting guilt tended to ``invert'' the evidence, thereby counterintuitively reducing their belief in the guilt of the accused. \citet{fernbach2011good} argue that the effect occurs because people give undue weight to the weak evidence and fail to take into account alternative evidence that more strongly favours the hypothesis at hand.

\paragraph{Implications for rule learning}
The weak evidence effect can be directly applied to rules: the evidence is represented by the rule antecedent; the  consequent corresponds to the outcome. The analyst can intuitively interpret each of the conditions in the antecedent as a piece of evidence in favour of the outcome. Typical of many machine learning problems is the uneven contribution of individual attributes to the prediction. Let us assume that the analyst is aware of the prediction strength of the individual attributes. 

If the analyst is to choose from a  rule containing only one strong condition (predictor) and another rule containing a strong predictor and a weak (weak enough to trigger this effect) predictor, according to the weak evidence effect would sway the analyst towards the shorter rule with one predictor.

\paragraph{Differentiation from information bias}
In the context of rule learning, the weak evidence effect needs to be contrasted with the information bias, which can lead to a preference for longer rules. The two phenomena influence the analysts at a different stage of working with rules. The information bias leads people to request more information, even if the additional information is not helpful. The weak evidence effect is triggered when the analyst evaluates a rule. If the rule contains a  condition that the analyst knows to be weak, this effect may cause this rule to be evaluated as less plausible than without the weak predictor.

Note that in Section~\ref{ss:information} we suggest that information bias may also apply when the user is readily  presented  with  more  information. In that case, we speculate that which bias will be triggered will depend on whether the strength of evidence is known. It is also possible that the effects of both biases will be combined.

\paragraph{Debiasing techniques}
\citet{martire2014interpretation} performed an empirical study aimed at evaluating what mode of communication of the strength of evidence is most resilient to the weak evidence effect. The surveyed modes of expression were numerical, verbal, a table, and a visual scale. It should be noted that the study was performed in the specific field of assessing evidence by a juror in a trial and the verbal expressions were following standards proposed by the Association of Forensic Science Providers \citep{willis2010standards}.\footnote{These  provide guidelines on the translation of numerical likelihood ratios into verbal formats. For example, likelihood ``$>1-10$'' is translated as ``weak or limited", and likelihood of ``$1000-10,000$'' as ``strong''.} The results clearly suggested that numerical expressions of evidence are most suitable for expressing uncertainty, which---in the scope of rule learning---is typically expressed through rule confidence.

Likelihood ratios studied by \citet{martire2014interpretation} are conceptually close to the lift metric, used to characterize association rules.  While lift is still typically presented as a number in machine learning user interfaces, there has been research towards communicating rule learning results in the natural language since at least 2005 \citep{strossa2005reporting}. With the recent resurgence of interest in interpretable models, the use of natural language has been taken up by commercial machine learning services, such as BigML, which allow to generate predictions via spoken questions and answers using Amazon Alexa voice service.\footnote{\url{https://bigml.com/tools/alexa-voice}} Similarly, machine learning interfaces increasingly rely on visualizations.
The research on debiasing of the weak evidence effect suggests that when conveying  machine learning results using modern means, such as transformation to natural language or through visualizations, care must be taken when numerical information is communicated. 
For example, some empirical evidence seems to suggest that it may not be advisable to replace the numerical strength of evidence with linguistic labels, which are frequently used in fuzzy systems  \cite{herrera2000approach} and also in rule learning.
It should be noted that the numerical representation is not without problems as well. In particular, the Steven's power law shows that relationship between the magnitude of a stimulus and its perceived magnitude is not linear \cite{kane1986stevens}. While this law was initially applied to physical stimuli, it has also been recently shown to hold for virtual concepts \cite{schulte2014stevens}.

\citet{martire2014interpretation} also observe a high level of miscommunication associated with low-strength verbal expressions. In these instances, it is ``appropriate to question whether expert opinions in the form of verbal likelihood ratios should be offered at all'' \citep{martire2014interpretation}. Transposing this result to the machine learning context, we suggest to consider an intentional omission of weak predictors from rules either directly by the rule learner or as part of feature selection.

\subsection{Unit Bias}
The unit bias refers to the tendency to give each unit similar weight while ignoring or underweighting the size of the unit \citep{geier2006unit}. 

\citet{geier2006unit} offered people various food items in two different sizes on different days and observed how this would affect the consumption of the food. They found that people ate a larger amount of food when the size of a single unit of the food item was big than when it was small. A possible explanation is that people ate one unit of food at a time without taking into account how big it was. Because the food was not consumed in larger amounts at any single occasion but was rather eaten intermittently, the behaviour led to higher consumption when a unit of food was larger.

\paragraph{Implications for rule learning}
Unit bias has been so far primarily studied for quite different purposes than is the domain of machine learning. Nevertheless, it can be very relevant to the domain of rule learning.

From a technical perspective, the number of conditions in rules is not important. What matters is the actual discriminatory power of the individual conditions, which can vary substantially.
However, following the application of unit bias, people can view conditions as units of similar importance, disregarding their sometimes vastly different discriminatory and predictive power.
 
We found some support for unit bias in the responses collected in \cite{furnkranz2019cognitive}:
``Rule one contains twice as many properties as rule 2 does for determining the edibility of a mushroom so that makes it statistically twice as plausible, hence much higher probability of being believable'', ``An extra group increases the likelihood.''
It follows from the responses above, that when assessing plausibility, these participants mainly relied on the count of conditions in the antecedent of the rule disregarding quality or predictive power of the conditions, which would correspond to the application of the unit bias. 

\paragraph{Debiasing techniques}

One of the common ways how regulators address unhealthy food consumption patterns related to varying sizes of packaging is  introduction of mandatory labelling of the size and calorie contents.
Following an analogy to clearly communicating the size of the food item, informing analysts about the discriminatory power of the individual conditions  may alleviate the unit bias. Such an indicator can be generated automatically, for example, by listing the number of instances in the entire dataset that meet the condition.

\section{Recommendations for Rule Learning Algorithms and Software}
\label{sec:recommendations}
This section provides a concise list of considerations that is aimed to raise awareness among machine learning practitioners regarding the availability of measures that could potentially suppress the effect of cognitive biases on the comprehension of rule-based models. We expect part of the list to be useful also for other symbolic machine learning models, such as decision trees. In our recommendations, we focus on systems that present the rule model to a human user.

\subsection{Adherence to conversational rules (maxims)}  
\label{ss:maxims}
Especially when results of machine learning are communicated to the general audience, it is essential to ensure that the automatically generated explanations of machine learning models do not violate the conversational rules set out by \citet{grice1975logic}. In table~\ref{tab:maxims} we provide definitions of the maxims adopted from \citet{thomas1992conversational} and link them to biases for which following the maxims could reduce the bias. 
Examples of recommendations following from the adherence to maxims for rule learning are given in the next subsection.

\begin{table}[h]
    \centering
\begin{tabular}{p{6cm}p{5.4cm}}
\toprule
     conversational maxim   & applicable cognitive biases\\
     \midrule
     \emph{Quantity}:   Make your contribution as informative  as is required (for the current purpose of the exchange). Do not make your contribution more informative than is required.  & unit bias, reiteration effect\\
     \emph{Quality}: Do not say what you believe to be false. Do not say that for which you lack adequate evidence.   & overconfidence and underconfidence, insensitivity to sample size, weak evidence effect\\
     \emph{Relation}: Be relevant.  & information bias, context and tradeoff contrast\\
     \emph{Manner}: Avoid obscurity of expression. Avoid ambiguity. Be brief (avoid unnecessary prolixity). Be orderly. & conjunction fallacy (misunderstanding of ``and''), ambiguity aversion, confusion of the inverse, primacy effect\\
     \bottomrule
\end{tabular}
    \caption{Conversational maxims (in wording from \cite{thomas1992conversational} with biases for which following the maxim could lead to reduction of the bias. Note that this alignment is only indicative, for example we always chose at most one maxim per  bias while multiple maxims could be fitting. }
    \label{tab:maxims}
\end{table}

\subsection{Representation of a rule}
The interpretation of natural language expressions used to describe a rule can lead to systematic distortions.
Our review revealed the following recommendations applicable to individual rules:
\begin{enumerate}
\item \textbf{Syntactic elements. }
There are several cognitive studies indicating that the conjunction AND is often misunderstood \citep{hertwig2008conjunction}, \citep[p. 95-96]{gigerenzer2001content}. The results of our experiments \citep{furnkranz2019cognitive} support the  conclusion  that AND needs to be presented unambiguously   in the rule learning context. Research has shown that AND ceases to be ambiguous when it is used to connect propositions  rather than categories.
To help avoid the confusion of the inverse effect, the communication of the implication construct IF THEN connecting antecedent and consequent should be made unambiguous.

Another important syntactic construct is negation (NOT). While processing of negation has not been included among the surveyed biases, our review of the literature (cf.\ Section~\ref{ss:beyond}) suggests that its use should be discouraged on the grounds that its processing requires more cognitive effort, and because the fact that a specific piece of information was negated may not be remembered in the long term.

\item \textbf{Conditions. } Attribute-value pairs (literals) comprising conditions are typically either formed of words with  semantics meaningful to  the user, or of codes that are not directly meaningful.
A number of biases can be triggered or strengthened by the (lack of) understanding of attributes and their values appearing in rules.  Providing easily accessible information on conditions in the rules, including their predictive power, can thus prove as an effective debiasing technique.
Redundant conditions and conditions with low relevance should not be included following the quantity conversational maxim.

When conditions contain words with negative valence,  these need to be reviewed carefully, since negative information is known to receive more attention and is associated with higher weight than positive information.

People have the tendency to put a higher emphasis on information they are exposed to first. By sorting the conditions by strength, machine learning software can conform to the manner conversational maxim (``be orderly''). The output could also visually delimit conditions in the rules based on their significance or predictive strength.

\item \textbf{Interestingness measures.} The values  should  be  communicated using numerical expressions. The use of alternate verbal expressions, with wordings such as ``strong relationship'' replacing specific numerical values, are discouraged
because
there is some evidence that such verbal expressions are prone to miscommunication. 

Currently, rule interest 
measures are typically represented as probabilities (confidence) or ratios (lift), whereas
results in cognitive science 
indicate that natural frequencies are better understood. 

The tendency of humans to ignore base rates and sample sizes (which closely relate to rule support) is a well-established fact in cognitive science. Results of our experiments on  inductively learned rules also provide evidence for this conclusion \citep{furnkranz2019cognitive}. Our proposition is that this effect can be addressed by presenting confidence (reliability) intervals  for the values of measures of interest, where applicable. 
\end{enumerate}

\subsection{Rule models}
In many cases, rules are not presented  in isolation to the analyst, but instead within a collection of rules comprising a rule model. Here, we relate the results of our review to the following aspects of rule models:

\begin{enumerate}
\setcounter{enumi}{3}
\item \textbf{Model size}.   An experiment by \citet{poursabzi2018manipulating} found that people are better able to simulate results of a smaller regression model composed of two coefficients than of a larger model composed of eight coefficients. The results indicate that removal of any unnecessary variables could improve model interpretability even though the experiment did not find a difference in the trust in the model based on the number of coefficients it consisted of. 
Similarly to regression models, rule models often incorporate output that is considered as marginally relevant. This can take a form of  (nearly) redundant rules or (nearly) redundant conditions in the rule.
Our analysis shows that such redundancies  can induce a number of biases, which may be accountable for misinterpretation of the model.
Size of a rule model can be reduced by utilizing various pruning techniques, or by using learning algorithms that allow the user to set or influence the size of the resulting model. Examples of such  approaches include those proposed by \citet{letham2015interpretable,lakkarajuinterpretable,wang2017bayesian}.  The  Interpretable Decision Sets  algorithm \citep{lakkarajuinterpretable} can additionally optimize for diversity and non-overlap of discovered rules,  directly countering the reiteration effect.

Eliminating redundancies can improve the adherence to the manner maxim (``be brief'') and the quantity conversational maxim. Following the quality maxim ``do not say for which you lack adequate evidence'', statistically insignificant  rules can be considered to be removed.
Another potentially effective approach to discarding some rules can be using domain knowledge or  constraints set by the user to remove the strong (e.g., highly confident), yet ``obvious'' rules confirming common knowledge.\footnote{For example, it is well-known that diastolic blood pressure rises with body mass index (DBP$\uparrow\uparrow$BMI). Rules confirming this relationship might under some circumstances be removed \citep{Kliegr:2011:SSA:2070639.2070657}.} Removal of weak rules could help to address the tradeoff contrast as well as the weak evidence effect.
Overall, eliminating redundancies and spurious rules can improve adherence to the quantity  conversational maxim (``do not say for which you lack evidence''). 
 
\item \textbf{Rule grouping.}
The rule learning literature has seen multiple attempts to develop methods for grouping  similar rules, often by clustering. Our review suggests that presenting clusters of similar rules can help to reduce cognitive biases caused by reiteration.

\item \textbf{Rule ordering.}
 Algorithms that learn rule lists provide mandatory ordering of rules, while the rule order in rule-set learning algorithms is not important. In either case, the rule order as presented to the user will affect perception of the model according to the conversational maxims and the primacy effect, among others. It is recommended to sort the presented rules by strength. However, due to the paucity of applicable research, it is unclear which particular  definition of rule strength would lead to the best results in terms of bias mitigation.
\end{enumerate}

\subsection{User Engagement}
Some results of our review suggest that increasing user interaction can help counter some biases. Several specific suggestions for machine learning user interfaces (UIs) follow:
\begin{enumerate}
\setcounter{enumi}{6}
\item \textbf{Domain knowledge}.
Selectively presenting  domain knowledge ``conflicting'' with the considered rule can help to invoke the 'consider-the-opposite' debiasing strategy.  Other research has shown that the plausibility of a model depends on compliance with monotonicity constraints 
\citep{freitas2014comprehensible}. 
We thus suggest that  UIs make background information on discovered rules easily accessible. 
\item \textbf{Eliciting rule annotation}.
Activating the deliberate ``System 2'' is one of the most widely  applicable debiasing strategies. One way to achieve this is to require accountability, e.g.,  through visual interfaces motivating users to annotate selected rules, which would induce the  'note-taking' debiasing strategy. For this technique to be effective, the created annotations would need to be consequently checked or used, either by a human or algorithmically. 
Giving people additional time to consider the problem has been in some cases shown as an effective debiasing strategy.  This can be achieved by making the selection process (at least) two-stage, allowing the user to  revise the selected rules.
 
 \item \textbf{User search for rules rather than scroll.}
Repeating rules can affect users via the mere exposure effect even if they are exposed to them even for a short moment, e.g., when scrolling a rule list. To mitigate this effect, the user interfaces should thus deploy alternatives to scrolling in discovered rules, such as search facilities. This can be combined with other measures for limiting the number of initially displayed rules. One option for improving diversity in the preview initially shown to the user is clustering the rules and showing a representative of each cluster. 
\end{enumerate}

\subsection{Bias inoculation}
In some  studies,  basic education   about specific biases, such as brief tutorials, decreased the fallacy rate.  This debiasing strategy has been called \emph{bias inoculation} in the literature.
\begin{enumerate}
\setcounter{enumi}{9}
    \item \textbf{Education.} 
Several studies have shown that providing explicit guidance and education on formal logic, hypothesis testing, and critical assessment of information can reduce fallacy rates in some tasks. However, the effect of psychoeducational methods is still a subject of dispute \cite{lilienfeld2009giving}, and cannot be thus recommended as a sole or sufficient measure.
\end{enumerate}

 \section{Limitations and Future Work}
 \label{sec:limitations}
Our goal was to examine whether cognitive biases can affect the interpretation of machine learning models and to propose possible remedies if they do. 
Since this field is untapped from the machine learning perspective, we tried to approach the problem holistically. Our work yielded a number of partial contributions, rather than a single profound result. We mapped applicable cognitive biases, identified prior works on their suppression, and proposed how these could be transferred to machine learning.

In the following, we outline some promising direction for future work.

\subsection{Validation through human-subject experiments}
\label{ss:validationfw}
All the identified shortcomings of the human judgment pertaining to the interpretation of inductively learned rules are based on empirical cognitive science research. For each cognitive bias, we 
provided a justification for how it would relate to machine learning. Due to the absence of applicable prior research in the intersection between cognitive science and machine learning, this justification is mostly based on authors' experience in machine learning. 

A critical next step is empirical validation of the effect of the selected cognitive biases. We have already described several user experiments aimed at validating selected cognitive biases in \citet{furnkranz2019cognitive}. Some other machine learning researchers have reported human-subject experiments that do not explicitly refer to cognitive biases, yet the cognitive phenomena they investigate may correspond to a known cognitive bias. One example is a study by \citet{narayanan2018humans}
, which investigated the effect of the number of cognitive chunks (conditions) in a rule on response time.
While the main outcome confirms the intuition  that higher complexity results in higher response times, this study has also revealed several unexpected patterns, such as that defining a new concept and reusing it leads to  a higher response time than repeating the description whenever that concept implicitly appears, even though this repetition means that subjects have to read more lines.
The findings could possibly be attributed to increased fluency due to repetition.

For several biases, we included examples based on data that we collected in our prior work \citep{furnkranz2019cognitive}. 
The main limitation of these examples is that they are a product of a user experiment performed with traditional psychological methods (questionnaires). Future work could use a complementary method -- observational study of users working in an operational environment (i.e. with real rule learning software).  To facilitate these efforts, it is necessary to adjust existing machine learning software with means for supporting cognitive experiments. In~\citet{ruleeditor}, we describe a rule editor adjusted for user studies of explainability, but similar software is needed also for other types of machine learning models.

Despite the existence of several early studies,  much more concentrated and systematic effort is needed to yield insights on the size of effect individual biases can have on the  understanding of intrinsically explainable machine learning models, such as rules and decision trees.

\subsection{Role of Domain Knowledge}
It has been long recognized that external knowledge plays an important rule in the rule learning process. Already \citet{mitchell1980need} recognized at least two distinct roles the external knowledge can play in machine learning: it can constrain the search for appropriate generalizations, and guide learning based on the intended use of the learned generalizations.
Interaction with domain knowledge has played an important role in multiple stages of the machine learning process. For example, it can improve semi-supervised learning \citep{carlson2010toward}, and in some applications, it is vital to convert discovered rules back into domain knowledge 
\citep[p.~288]{jf:Book-Nada}. Some results also confirm the common intuition that compliance to constraints valid in the given domain increases the plausibility of the learned models \citep{freitas2014comprehensible}. 

Our review shows that domain knowledge can be one of the important instruments in the toolbox aimed at debiasing interpretation of discovered rules. To give a specific example, the presence or strength of the reiteration effect depends on the familiarity of the subject with the topic area from which the information originates \citep{boehm1994validity} (cf. also Section~\ref{ss:reiteration}). Future work should focus on  a systematic review of the role of domain knowledge on activation or inhibition of cognitive phenomena applicable to the interpretability of rule learning results.

\subsection{Individual Differences}
The presence of multiple cognitive biases and their strength have been linked to specific personality traits. For example,  overconfidence and the rate of conjunctive fallacy  have been shown to be inversely related to  numeracy \citep{winman2014role}. According to \citet{juslin2011reducing}, the application of the averaging heuristic rather than the normative multiplication of probabilities seems to depend on the working memory capacity and/or motivation.

Some research can even be interpreted as indicating that data analysts can be more susceptible to the confirmation bias  than the general population. An experiment reported by \citet{wolfe2008locus} shows that subjects who defined good arguments as those that can be ‘‘proved by facts’’ (this stance, we assume, would also apply to many data analysts) were more prone to exhibiting the confirmation bias.\footnote{This tendency is explained by \citet{wolfe2008locus} as follows: ``For people with this belief, facts and support are treated uncritically. \ldots More importantly, arguments and information that may support another side are not part of the schema and are also ignored.''}  \citet{stanovich2013myside} show that the incidence of confirmation bias is surprisingly not related to general intelligence. This suggests that even highly intelligent analysts can be affected. 
\citet{albarracin2004role} propose that the susceptibility to the confirmation bias can depend on one's personality traits. They also present a diagnostic tool called ``defense confidence scale'' that can identify individuals who are prone to confirmational strategies. 
While it has been shown that education on some biases mitigates their effect, it would not be practical and cost-effective to educate all end-users.
Further research into  personality traits of users of machine learning outputs, as well as into the development of appropriate personality tests, would help to better target education focused on debiasing.

\subsection{Effectiveness and negative effects of debiasing techniques}
Debiasing strategies discussed in this article aim at mitigating negative effects by eliminating or reducing the biases. While there are very few empirical studies on the incidence of biases in the context of rule learning, there is almost no research on possible negative effects of debiasing strategies. For this reason, as a first step before deploying a debiasing strategy, we suggest that it should be examined that the actually targeted bias occurs and can distort decision making significantly enough to warrant the application of a debiasing strategy. Some early research \cite{pagliaro2020cognitive} suggests that application of a debiasing strategy can in some cases distort decision making even more than if it is left untreated. For this reason, we also suggest that it should be verified that the applied debiasing strategy provides a sufficient improvement over the baseline before it is deployed.   Future research should thus focus on understanding the trade-off between the costs and feasibility of a particular debiasing technique and its effectiveness.

\subsection{Extending Scope Beyond Biases}
\label{ss:beyond}
There is a number of cognitive phenomena affecting the interpretability of rules, which are not classified as cognitive biases.
Remarkably, since 1960 there is a consistent line of work by psychologists 
studying cognitive processes related to rule induction, which is centred around the so-called \emph{Wason's 2-4-6 problem} \citep{wason1960failure}. Cognitive science research on rule induction in humans has so far not been noticed in the rule learning subfield of machine learning.\footnote{Based on our analysis of cited reference search in Google Scholar for \citep{wason1960failure}.} It was out of the scope of the objectives of this review to conduct an analysis of the significance of these results 
for rule learning, nevertheless, we believe that such investigation could bring interesting insights for a cognitively-inspired design of rule learning algorithms.

Another promising direction for further work is research focused on the interpretation of negations (``not'').
Experiments conducted by \citet{jiang2014affective}  show that the mental processes involved in processing  negations slow down reasoning. Negation can be also sometimes ignored or forgotten \citep{deutsch2009fast}, as it decreases veracity of long-term correct remembrance of  information. 

Most rule learning algorithms are capable of generating rules containing negated literals. For example, a healthy company  can be represented as \texttt{status = not(bankrupt)}.
Our precautionary suggestion based on the interpretation of results obtained in general studies performed in experimental psychology \citep{deutsch2009fast} and neurolinguistics \citep{jiang2014affective} is that  the use of negation in the discovered rules that are to be presented to the user should be very carefully considered..
Due to the adverse implications of the use of negation on cognitive load and remembrance, empirical research focused on the interpretability of negation in machine learning is urgently needed.

\subsection{Extending Scope Beyond Inductively Learned Rules}
\label{ss:beyond-rules}

Most of our discussion focused on logical models in the form of symbolic rules. However, the underlying principle is more general: for the automatic generation of arbitrary but human-interpretable models, we need to understand how humans perceive these models. Findings from the cognitive sciences, such as those regarding cognitive biases, may help to formulate models in a better and more convincing way, not only for logical models, but also for visual, structural, mathematical, or probabilistic models. The principles should not only apply to inductively learned models, but also to models that have been obtained by other forms of reasoning, such as deduction, search-based planning, or Case-Based Reasoning. However, further work will be necessary to understand the specific requirements of each of these cases, such as whether the model is guaranteed to be correct (deduction) or that the model consists of one or more suitably chosen prior experiences as in Case-Based Reasoning. Also, different users may have different levels of expertise and might require different levels of abstraction. For example, the acceptance of mathematical models will certainly depend on the user's proficiency in mathematics, and similar cases can be made for domain experts in various application areas such as medicine. While we have provided some pointers to work that links specific biases to the user's background (such as formal training in statistics, intelligence), more work in this direction is needed.

\section{Conclusion}
To our knowledge, cognitive biases have not yet been systematically discussed in relation to the interpretability of rule learning results.  We thus initiated this review of research published in cognitive science with the intent of providing a psychological basis to further research in inductive rule learning algorithms, and to the way their results are communicated.  Our review covered twenty cognitive biases, heuristics, and effects that can give rise to  systematic errors when inductively learned rules are interpreted. 

For most biases and heuristics included in our review, psychologists have proposed ``debiasing'' measures. 
Application of prior empirical results obtained in cognitive science allowed us to propose several methods that could be effective in suppressing these cognitive phenomena when machine learning models are interpreted.
For example, the proposed representation of discovered rules based on ``frequency formats'' of \cite{gigerenzer1995improve} is hypothesized to reduce the number of judgmental errors attributable to the empirically proven tendency of users to overvalue rule confidence at the expense of rule support (insensitivity to sample size), as well to 
 the  misunderstanding of the \emph{and} conjunction.

Even with nearly two hundred articles included in our review, we processed only a fraction of potentially relevant psychological studies of cognitive biases; yet, we were able to find only very limited research dealing specifically with cognitive biases in machine learning. Future work should particularly focus on empirical evaluation of the effects of cognitive biases in this domain.

 \subsection*{\bf Acknowledgments}
 \noindent{\small
 TK was supported by long term institutional support of research activities. \v{S}B and TK were supported by grant IGA 33/2018 by Faculty of Informatics and Statistics, Prague University of Economics and Business. 
 The contribution of TK was partially supported by the EU H2020 programme, grant No 857446 HeartBIT\_4.0. An initial version of this review was published as a part of TK's PhD thesis at Queen Mary University of London.

We are grateful for the helpful comments of the reviewers of this paper.
}

\bibliography{bibliography}

\end{document}